\documentclass[conference]{IEEEtran}
\IEEEoverridecommandlockouts

\usepackage{flushend}
\usepackage{tabularx}
\usepackage{tablefootnote}
\usepackage[normalem]{ulem}
\usepackage[switch]{lineno}
\usepackage{cite}
\usepackage{amsmath,amssymb,amsfonts}
\usepackage{algorithmic}
\usepackage{graphicx}
\usepackage{textcomp}
\usepackage{xcolor}
\def\BibTeX{{\rm B\kern-.05em{\sc i\kern-.025em b}\kern-.08em T\kern-.1667em\lower.7ex\hbox{E}\kern-.125emX}}
\usepackage{booktabs}
\usepackage{multirow}
\usepackage{array}
\usepackage{url}
\usepackage{hyperref}
\usepackage{orcidlink}
\newcolumntype{L}[1]{>{\raggedright\arraybackslash}p{#1}}

\usepackage{enumitem}
\usepackage{tcolorbox} \tcbuselibrary{skins,breakable,theorems}
\tcbuselibrary{theorems}

\usepackage{array}
\usepackage{multirow}

\usepackage[dvipsnames]{xcolor}

\newcolumntype{L}[1]{>{\raggedright\arraybackslash}p{#1}}
  
\usepackage{tikz}
\usetikzlibrary{shapes.geometric, arrows.meta, positioning, fit, backgrounds, calc, shadows.blur, decorations.pathreplacing, decorations.markings}
\usetikzlibrary{arrows.meta, decorations.markings, calc, positioning, shapes.geometric}

\definecolor{rediblue}{HTML}{2962FF}
\definecolor{redigreen}{HTML}{00897B}
\definecolor{redipurple}{HTML}{7B1FA2}
\definecolor{redigray}{HTML}{546E7A}
\definecolor{rediorange}{HTML}{EF6C00}
\definecolor{darktext}{HTML}{212121}
\definecolor{rapscolor}{HTML}{C62828}

\definecolor{rediBlue}{HTML}{2563EB}
\definecolor{rediBlueLight}{HTML}{DBEAFE}
\definecolor{aidrinAmber}{HTML}{D97706}
\definecolor{aidrinAmberLight}{HTML}{FEF3C7}
\definecolor{setgoGreen}{HTML}{059669}
\definecolor{setgoGreenLight}{HTML}{D1FAE5}
\definecolor{rawRed}{HTML}{DC2626}
\definecolor{catalogPurple}{HTML}{7C3AED}
\definecolor{bgGray}{HTML}{F8FAFC}
\definecolor{textDark}{HTML}{1E293B}
\definecolor{textMid}{HTML}{475569}
\definecolor{arcGray}{HTML}{CBD5E1}
\definecolor{assessSlate}{HTML}{64748B}

\begin{document}

\newlist{henum}{enumerate}{1}
\setlist[henum,1]{label=(H\arabic*)}

\newtcolorbox[auto counter]{finding}[1][]{float=!htbp,
    colback=blue!5,           colframe=blue!40,         boxrule=0pt,              leftrule=2mm,             sharp corners,            before upper={\textbf{Finding~\thetcbcounter:}~}, fontupper=\normalfont,    }

\title{Automated Data Readiness for Scientific AI}

\author{\IEEEauthorblockN{Sean R. Wilkinson\,\orcidlink{0000-0002-1443-7479}, Valentine G. Anantharaj\,\orcidlink{0000-0002-9356-1311}, Jong Youl Choi\,\orcidlink{0000-0002-6459-6152}, Ketan Maheshwari\,\orcidlink{0000-0003-3800-662X},\\
Marshall McDonnell\,\orcidlink{0000-0002-3713-2117}, Massimiliano Lupo Pasini\,\orcidlink{0000-0002-4980-6924}, Polina Shpilker\,\orcidlink{0000-0002-6761-7326}, Renan Souza\,\orcidlink{0000-0002-1794-808X},\\
Patrick Widener\,\orcidlink{0000-0002-5882-0816}, Sarp Oral\,\orcidlink{0000-0001-8745-7078}, and Wesley Brewer\,\orcidlink{0000-0002-3639-3956}}\IEEEauthorblockA{Oak Ridge National Laboratory, Oak Ridge, TN, USA}
\thanks{Corresponding author: Sean R. Wilkinson (wilkinsonsr@ornl.gov).}}

\maketitle

\begin{abstract}
Leadership computing facilities steward large-scale scientific datasets that routinely require substantial transformation before serving as AI training data. However, no existing framework fully unifies automated transformation, readiness assessment, provenance tracking, and agent-native deployment. We present REDI, an open-source framework that addresses this gap through a unified five-stage pipeline (ingest, preprocess, transform, structure, and output) with per-stage instrumentation for reproducibility and deployment as an agent-callable skill; companion tool SetGo automates FAIR compliance and catalog publication. Evaluated across climate, proteomics, materials science, and nuclear fusion, REDI transforms all datasets from raw to AI-ready, with outputs validated against domain-expert references, and preliminary results show near-ideal parallel scaling to 100 nodes on Frontier for the climate case. Provenance-instrumented profiling reveals file I/O as the dominant pipeline cost, with format selection a first-order optimization lever. These results establish REDI as a cross-domain platform providing automated data readiness for scientific AI, transforming data preparation bottlenecks into reproducible, reusable community assets.
\end{abstract}
 

\section{Introduction}
\label{sec:intro}

Artificial Intelligence (AI) is increasingly recognized as a transformative capability for scientific discovery, in part due to its ability to support data-driven modeling, surrogate simulation, uncertainty quantification, and the development of foundation models across a growing range of scientific domains. Augmenting traditional simulation and experimental workflows with learning-based approaches has the potential to accelerate discovery across scales and modalities. 

Realizing this potential, however, depends critically on the availability of computable scientific data. Specifically, it requires datasets that are not only archived or preserved but also structured, validated, and semantically enriched such that they can be directly consumed by large-scale AI workflows. As scientific instruments and experiments produce ever-growing volumes of complex, heterogeneous data, the transition from raw data to AI-ready assets has emerged as a new and challenging component of the scientific data lifecycle.

To this end, we present the Readiness Engine for Data Integration (REDI),\footnote{REDI source code: \url{https://doi.org/10.11578/dc.20260702.1}} an open-source framework that automates AI data readiness
through a unified five-stage pipeline, namely ingest, preprocess, transform, structure, and
output. To our knowledge, REDI is the first framework to integrate automated
transformation, readiness assessment, provenance tracking, and validation within
a lifecycle-aware architecture for scientific computing.

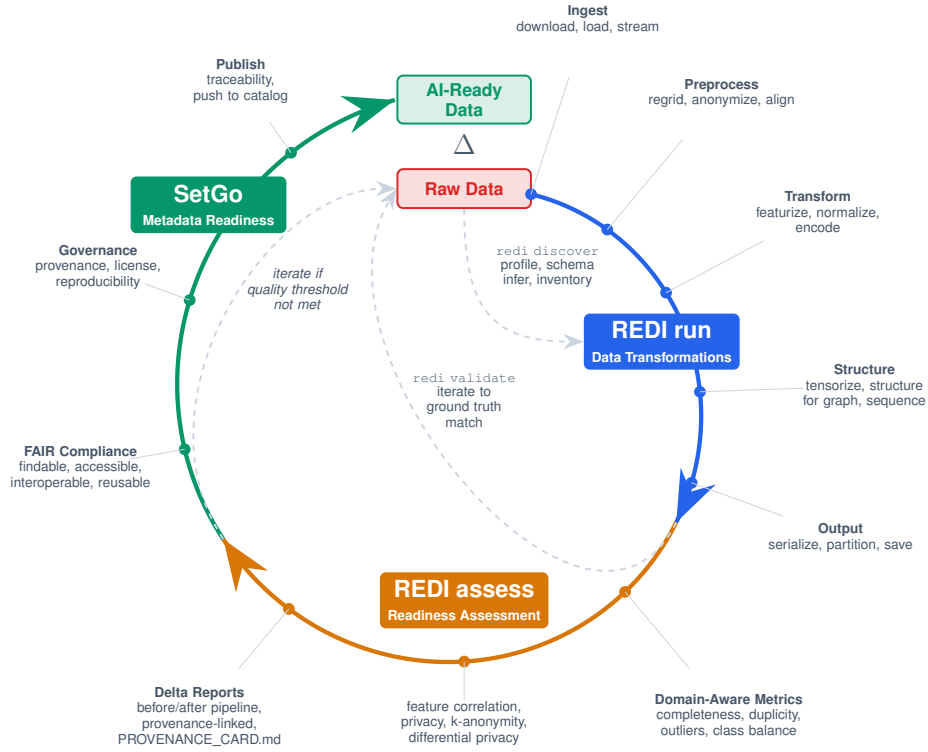
\begin{figure*}
    \centering
    \resizebox{0.7\textwidth}{!}{\begin{tikzpicture}[
    font=\sffamily,
    every node/.style={text=textDark},
    stage/.style={
        font=\sffamily\small\bfseries,
        rounded corners=3pt,
        minimum width=2.2cm,
        minimum height=0.7cm,
        inner sep=4pt,
        align=center,
    },
    substage/.style={
        font=\sffamily\scriptsize,
        text=textMid,
        align=center,
    },
]

\def\Rstart{3.8}
\def\Rend{5.6}
\def\Rmid{4.7}
\def\Rlabel{7.2}

\draw[
    color=rediBlue, line width=2.2pt,
    postaction={decorate,
        decoration={markings,
            mark=at position 1 with {\arrow[rediBlue,scale=2]{Stealth}}
        }
    }
]
plot[domain=90:-30, samples=80, variable=\t]
    ({(\Rstart + (\Rend-\Rstart)*(90-\t)/360)*cos(\t)},
     {(\Rstart + (\Rend-\Rstart)*(90-\t)/360)*sin(\t)});

\draw[
    color=aidrinAmber, line width=2.2pt,
    postaction={decorate,
        decoration={markings,
            mark=at position 1 with {\arrow[aidrinAmber,scale=2]{Stealth}}
        }
    }
]
plot[domain=-30:-150, samples=80, variable=\t]
    ({(\Rstart + (\Rend-\Rstart)*(90-\t)/360)*cos(\t)},
     {(\Rstart + (\Rend-\Rstart)*(90-\t)/360)*sin(\t)});

\draw[
    color=setgoGreen, line width=2.2pt,
    postaction={decorate,
        decoration={markings,
            mark=at position 1 with {\arrow[setgoGreen,scale=2]{Stealth}}
        }
    }
]
plot[domain=-150:-257, samples=80, variable=\t]
    ({(\Rstart + (\Rend-\Rstart)*(90-\t)/360)*cos(\t)},
     {(\Rstart + (\Rend-\Rstart)*(90-\t)/360)*sin(\t)});

\pgfmathsetmacro{\rRedi}{\Rstart + (\Rend-\Rstart)*60/360}
\node[stage, fill=rediBlue, text=white, minimum width=2.8cm]
    at ({(\rRedi)*cos(30)}, {(\rRedi)*sin(30)-1.0})
    {\large REDI run\\[2pt]\scriptsize Data Transformations};

\pgfmathsetmacro{\rAidrin}{\Rstart + (\Rend-\Rstart)*180/360}
\node[stage, fill=aidrinAmber, text=white, minimum width=2.8cm]
    at ({(\rAidrin-0.9)*cos(-90)}, {(\rAidrin-0.9)*sin(-90)+0.2})
    {\large REDI assess\\[2pt]\scriptsize Readiness Assessment};

\pgfmathsetmacro{\rSetgo}{\Rstart + (\Rend-\Rstart)*300/360}
\node[stage, fill=setgoGreen, text=white, minimum width=2.8cm]
    at ({(\rSetgo)*cos(-212)-0.1}, {(\rSetgo)*sin(-212)+0.7})
    {\large SetGo\\[2pt]\scriptsize Metadata Readiness};

\node[
    stage,
    fill=rawRed!15, draw=rawRed, line width=1pt, text=rawRed,
    minimum width=2.4cm,
] (raw) at (0, \Rstart) {Raw Data};

\node[font=\sffamily\Large\bfseries, text=textMid]
    at (0, {(\Rstart + \Rend*sin(90)-0.2)/2}) {$\Delta$};

\node[substage, text width=2.4cm] (discover) at (1.5, \Rstart-1.4)
    {\texttt{redi discover}\\[-1pt]profile, schema\\infer, inventory};

\draw[
    arcGray, dashed, line width=1.0pt,
    -{Stealth[scale=1.4, color=arcGray]},
]
    (raw.south)
    .. controls (0, \Rstart-2.2) and (0.5, {\rRedi*sin(30)-1.0}) ..
    ({\rRedi*cos(30)-1.4}, {\rRedi*sin(30)-1.0});
    
\node[
    stage,
    fill=setgoGreen!15, draw=setgoGreen, line width=1pt, text=setgoGreen,
    minimum width=2.4cm,
] (ready) at ({\Rend*cos(90)}, {\Rend*sin(90)-0.2}) {AI-Ready\\Data};

\draw[
    arcGray, dashed, line width=1.0pt,
    -{Stealth[scale=1.4, color=arcGray]},
]
    (3.81, -2.20)
    .. controls (2.26, -5.34) and (-3.00, 0.80) ..
    (raw.west);

\node[substage, text width=2.4cm, text=textMid]
    at (0.0, 0.0)
    {\texttt{redi validate}\\[-1pt]iterate to\\ground truth\\match};

\draw[
    arcGray, dashed, line width=1.0pt,
    -{Stealth[scale=1.4, color=arcGray]},
]
    (-4.33, -2.50)
    .. controls (-5.98, 0.01) and (-3.50, 3.80) ..
    (raw.west);

\node[substage, text width=2.4cm, text=textMid]
    at (-3.0, 2.0)
    {\textit{iterate if\\quality threshold\\not met}};

\foreach \a/\lbl/\sub in {
    72/{Ingest}/{download, load, stream},
    50/{Preprocess}/{regrid, anonymize, align},
    28/{Transform}/{featurize, normalize,\\encode},
    2/{Structure}/{tensorize, structure\\for graph, sequence},
    -20/{Output}/{serialize, partition, save}
}{
    \pgfmathsetmacro{\rdot}{\Rstart + (\Rend-\Rstart)*(90-\a)/360}
    \fill[rediBlue] ({\rdot*cos(\a)}, {\rdot*sin(\a)}) circle (3pt);
    \node[substage, text width=2.8cm] at ({\Rlabel*cos(\a)}, {\Rlabel*sin(\a)})
        {\textbf{\lbl}\\\sub};
    \draw[arcGray, line width=0.4pt]
        ({\rdot*cos(\a)}, {\rdot*sin(\a)}) --
        ({(\Rlabel-1.1)*cos(\a)}, {(\Rlabel-1.1)*sin(\a)});
}

\foreach \a/\lbl/\sub in {
    -50/{\textbf{Domain-Aware Metrics}}/{completeness, duplicity,\\outliers, class balance},
    -130/{\textbf{Delta Reports}}/{before/after pipeline,\\provenance-linked,\\PROVENANCE\_CARD.md}
}{
    \pgfmathsetmacro{\rdot}{\Rstart + (\Rend-\Rstart)*(90-\a)/360}
    \fill[aidrinAmber] ({\rdot*cos(\a)}, {\rdot*sin(\a)}) circle (3pt);
    \node[substage, text width=3cm] at ({(\Rlabel+0.2)*cos(\a)}, {(\Rlabel+0.2)*sin(\a)})
        {\lbl\\\sub};
    \draw[arcGray, line width=0.4pt]
        ({\rdot*cos(\a)}, {\rdot*sin(\a)}) --
        ({(\Rlabel-0.9)*cos(\a)}, {(\Rlabel-0.9)*sin(\a)});
}

\pgfmathsetmacro{\myshift}{1.6}
\pgfmathsetmacro{\rdot}{\Rstart + (\Rend-\Rstart)*(90-(-90))/360}
\fill[aidrinAmber] ({\rdot*cos(-90)}, {\rdot*sin(-90)}) circle (3pt);
\node[substage, text width=3cm] at ({(\Rlabel+0.2)*cos(-90)}, {(\Rlabel+0.2)*sin(-90)+\myshift})
    {feature correlation,\\privacy, k-anonymity,\\differential privacy};
\draw[arcGray, line width=0.4pt]
    ({\rdot*cos(-90)}, {\rdot*sin(-90)}) --
    ({\rdot*cos(-90)}, {(\Rlabel+0.2)*sin(-90)+\myshift+0.4});

\foreach \a/\lbl/\sub in {
    -170/{\textbf{FAIR Compliance}}/{findable, accessible,\\interoperable, reusable},
    -200/{\textbf{Governance}}/{provenance, license,\\reproducibility},
    -235/{\textbf{Publish}}/{traceability,\\push to catalog}
}{
    \pgfmathsetmacro{\rdot}{\Rstart + (\Rend-\Rstart)*(90-\a)/360}
    \fill[setgoGreen] ({\rdot*cos(\a)}, {\rdot*sin(\a)}) circle (3pt);
    \node[substage, text width=2.8cm] at ({(\Rlabel-0.2)*cos(\a)}, {(\Rlabel-0.2)*sin(\a)})
        {\lbl\\\sub};
    \draw[arcGray, line width=0.4pt]
        ({\rdot*cos(\a)}, {\rdot*sin(\a)}) --
        ({(\Rlabel-1.1)*cos(\a)}, {(\Rlabel-1.1)*sin(\a)});
}

\end{tikzpicture}

 }
    \caption{The REDI-SetGo iterative data readiness lifecycle. Raw data goes through a sequence of transformations via REDI, followed by domain-aware readiness assessment via the built-in \texttt{redi assess} command. Finally, metadata analysis and publishing to catalogs such as CKAN (Comprehensive Knowledge Archive Network), Hugging Face, and OpenMetadata are handled by SetGo. The ``$\Delta$'' denotes the readiness improvement from raw to AI-ready; several refinement iterations may occur before the AI-ready dataset is published.}
    \label{fig:lifecycle}
\end{figure*}

\subsection{The Data Readiness Challenge}

The term \emph{data readiness} is inherently context-dependent because different computational workflows impose distinct requirements on data structure, semantics, and validation. It is therefore necessary to ask, ``data readiness for what?'' In this work, we primarily focus on readiness with respect to preparing data for training scientific AI models, particularly foundation models~\cite{bommasani2021opportunities}. 
Within this scope, we define \emph{AI-ready data} as machine-readable datasets that (a) have undergone domain-appropriate cleaning, validation, and feature engineering, and (b) include sufficient metadata to support reproducible model training and evaluation.

Scientific data management has long emphasized stewardship across the stages of the data lifecycle, including acquisition, curation, preservation, access, and reuse. Frameworks that abide by the FAIR data principles~\cite{wilkinson2016fair} have provided essential guidance for making data findable, accessible, interoperable, and reusable. Although FAIR compliance is a necessary foundation, it is not sufficient for AI-driven science~\cite{clark2024}.  FAIR data may still be unsuitable for AI/ML uses due to missing structure, inconsistent representations, insufficient metadata for feature extraction, or misalignment with model training requirements. Conversely, data that are already used for bespoke training may still lack the necessary metadata for broader reuse. Our work addresses these issues through cooperative tools: REDI manages data transformations while SetGo~\cite{wilkinson2026setgo} ensures metadata readiness.

Data Readiness for AI (DRAI)~\cite{brewer2026datareadiness} extends traditional data quality and stewardship concepts by explicitly addressing the downstream computational and statistical requirements of AI workflows. This includes readiness with respect to schema consistency, feature completeness, provenance traceability, governance constraints, and validation against domain expectations. In the absence of standardized readiness definitions and processes, scientific teams frequently develop ad hoc preprocessing pipelines that are difficult to reproduce, audit, or generalize across projects and facilities.

\subsection{Cost and Complexity of Data Preparation}

The lack of standardized, automated readiness workflows has significant practical consequences. Practitioners developing AI models consistently report that 70\%--80\% of project effort is devoted to preparing data for training and inference, instead of model development and scientific interpretation~\cite{Lawrence2017}. A complementary problem arises when published data already appears AI-ready: in the absence of standardized workflows, preparation steps are often undocumented, making results difficult to reproduce or audit. REDI addresses both deficiencies by automating preparation when it is absent and providing provenance and transparency when it has already occurred.
Notably, in 2026, the National Academies of Sciences, Engineering, and Medicine (NASEM) identified data preparation and cleaning as ``most ripe for AI acceleration''~\cite{nasem2026frontiers}, yet the computational cost of preprocessing scientific data at leadership scale remains largely uncharacterized. This paper directly addresses that gap through empirical profiling. Moreover, when preprocessing costs are characterized, the dominant bottlenecks vary significantly by domain and data regime, with direct implications for optimization at leadership scale.

From a data lifecycle perspective, this preprocessing burden represents a critical friction point between data production and data reuse. Every new AI application often reconstructs similar preparation steps, and this can lead to duplicated effort, inconsistent assumptions, and reduced trust in derived results. 
Recognizing this problem, the authors of the NASEM report call for ``unified data ecosystems where findable, shared, and version-controlled repositories are kept with standardized preprocessing procedures and in-depth metadata''~\cite{nasem2026frontiers}.

Data lifecycles are complex and nuanced. Data are modified, shared, branched, reused, and repurposed, and each such action implies the need for a reconsideration of data preparation. 
Human operators, human-written code, and---increasingly---LLM-generated code can all introduce errors, inconsistencies, and long-lived, hard-to-find problems. Reuse and repurposing of data can expose unanticipated side effects in data lifecycle maintenance. Increased collaboration and wider data availability combine with these factors to make data readiness a cyclic process. This is reflected in the iterative data flow through the REDI framework (Fig.~\ref{fig:lifecycle}), in which a dataset is reconsidered if a sufficient readiness quality threshold is not met. The eventual output of this process is a refined AI-ready dataset that can be published for general use and potentially become a basis for new data combinations that are also candidates for refinement by REDI. This cyclical nature further motivates the development of standardized, lifecycle-aware, and automatable data readiness pipelines that can bridge the gap between FAIR data stewardship and operational AI deployment.

\subsection{Contributions of This Work}

The central contribution of this work is the design and implementation of REDI
as a cross-domain, lifecycle-aware framework for automated data transformation
and readiness. Additional details are provided below.

\begin{itemize}[label={},leftmargin=0pt]

  \item \textbf{A unified, cross-domain data readiness framework.} Although 
  frameworks exist for domain-specific data preparation (e.g.,  PhysicsNeMo-Curator for computational fluid dynamics (CFD)~\cite{chandrasekar2023physicsnemo} and Anemoi-datasets for climate science~\cite{anemoi_datasets}), no existing
  general-purpose framework provides a unified, operational model for
  data readiness across heterogeneous scientific domains and computational
  workflows. REDI addresses this gap by enabling consistent and interpretable
  readiness assessment, reusability of common preprocessing tasks, and a shared
  preprocessing infrastructure for nascent multi-domain foundation
  models~\cite{xia2025nature, menon2026scientific, soares2025towards}.

  \item \textbf{Domain-aware, automated transformation.} REDI dynamically
  determines preprocessing actions based on detected data types, formats, and
  scientific context, including personally identifiable information (PII)
  anonymization for bioinformatics datasets,
  regridding and coordinate standardization for climate data, and graph-based
  encoding for materials modeling, among others. Outputs are produced in
  standardized, interoperable formats (Zarr, NPZ,
  ADIOS) to support FAIR principles.

  \item \textbf{Agentic integration.} While REDI is a standalone framework, it also functions 
  as a data readiness skill within agentic programming environments such as Claude Code and OpenAI Codex to enable automated pipeline construction and execution with full provenance, versioning, and reproducibility.

  \item \textbf{An end-to-end data-for-AI factory.} REDI operates as the
  transformation core of a complete data readiness lifecycle and is complemented by
  a built-in \texttt{redi assess} command for domain-aware quality assessment
  and SetGo for metadata readiness, FAIR compliance validation, and automated
  catalog publishing.

  \item \textbf{Cross-domain empirical validation.} We evaluate REDI against
  bespoke, domain-specific preprocessing pipelines across climate science
  (ClimaX), proteomics (OpenFold), materials science (HydraGNN), and nuclear fusion (XGC1). 
  Across all cases, REDI achieves readiness-equivalent outputs while significantly reducing
  preprocessing complexity and manual engineering effort.

\end{itemize}

The rest of this paper is structured as follows: Section~\ref{sec:background} describes related work across the topic areas that motivate REDI; Section~\ref{sec:methods} describes REDI's architecture, modes of operation, and integration with SetGo; Section~\ref{sec:usecases} introduces the scientific domains and datasets used to evaluate REDI; Section~\ref{sec:evaluation} presents empirical results across discover mode behavior, readiness assessment, pipeline performance, and preliminary parallel scalability; and Section~\ref{sec:conclusions} concludes with broader findings, limitations, and future work.

\section{Background}
\label{sec:background}

Four intersecting concerns motivate REDI: the readiness vocabulary needed to characterize the progression from raw to AI-ready data; the HPC storage characteristics that constrain pipeline design at leadership scale; the fragmented tool landscape that addresses this lifecycle only in isolation; and the novel failure modes introduced by agentic AI workflows without structured readiness guidance. The subsections below survey each area.

\subsection{Scientific Data Lifecycle and Readiness Levels}

User facilities and leadership computing centers of the United States Department of Energy (DOE) face stringent requirements for provenance, validation, governance, and long-term stewardship, given the scale of scientific data production and the public accountability inherent in federally funded research~\cite{doepublicaccessplan}. The FAIR data principles~\cite{wilkinson2016fair, wilkinson2025fairworkflows} have been widely adopted across DOE programs and facilities; FAIR compliance alone is insufficient for AI readiness, however: a dataset can be fully FAIR yet still require substantial normalization, format conversion, or feature engineering before it can serve as training data.

Unlike single-task pipelines, cross-domain pretraining for scientific foundation models~\cite{nguyen2023climax, ahdritz2024openfold, xia2025nature, menon2026scientific, wang2024orbit} demands consistent preprocessing conventions: differing normalization schemes, coordinate systems, or feature encodings are silently inherited during training and introduce subtle inconsistencies into learned representations that are difficult to detect post hoc. A shared preprocessing infrastructure enforcing consistent readiness operations across domains is therefore not merely convenient but structurally required.

Characterizing this raw-to-AI-ready progression requires explicit readiness taxonomies. Lawrence~\cite{Lawrence2017} introduced the foundational data readiness levels by organizing datasets into Band~C (accessible), Band~B (usable), and Band~A (useful for a specific purpose). This taxonomy has since been extended and synthesized across communities~\cite{hiniduma2025data}. These taxonomies share a common limitation: they are descriptive rather than operational, identifying a readiness state without providing an automatable path for advancing it.
The Data Readiness for AI (DRAI) construct~\cite{brewer2026datareadiness} maps each band transition onto a canonical IPTSO (Ingest, Preprocess, Transform, Structure, Output) pipeline stage, and defines five quantitative readiness levels from Level~1 (RAW) through Level~5 (FULLY\_AI\_READY), thus enabling reproducible quantification of readiness improvement.

\subsection{HPC Parallel Storage and Scientific File Formats}

Preprocessing at leadership scale commonly runs on parallel filesystems such as Lustre, which stripe data across many storage targets for high aggregate bandwidth but route each file operation (open, stat, close) through a metadata server; with hundreds of thousands of per-sample files, this metadata path can dominate regardless of available bandwidth~\cite{george2025lustre}. Object-based stores such as DAOS shift this profile by distributing metadata differently, but the general lesson holds: many-small-file access patterns, not raw bandwidth, tend to bound preprocessing at scale.

Scientific HPC datasets are stored in formats that encode domain-specific structural
assumptions. HDF5~\cite{folk2011hdf5} is a general-purpose hierarchical container supporting chunked
storage, compression, and random access; NetCDF-4~\cite{rew1990netcdf} builds on HDF5 and is the standard
format for atmospheric and climate data. ADIOS2~\cite{godoy2020adios} is a
streaming I/O library designed for high-throughput in situ analytics and simulation I/O. Zarr~\cite{moore2023zarr}
stores chunked N-dimensional arrays in a directory hierarchy that supports parallel
writes without file locking, making it well suited to multi-worker preprocessing. By
contrast, NPZ (NumPy's ZIP-based archive) and LMDB~\cite{chu2011lmdb} (a memory-mapped key-value store
widely used in machine learning (ML) datasets) do not support parallel random access and impose
significant per-entry overhead on Lustre metadata servers when per-sample sharding spans large datasets. A preprocessing framework for leadership-scale HPC must therefore: (a)~support
native I/O for HDF5, NetCDF-4, ADIOS2, and Zarr; (b)~select output formats matched
to the target training regime; and (c)~distribute file operations across parallel
workers to match leadership-class dataset scales.

\subsection{Existing Preprocessing Frameworks and Tools}

The existing transformation and readiness landscape reveals pervasive fragmentation. General-purpose data quality platforms such as Great Expectations~\cite{gong2021great} detect and report quality problems but do not resolve them, providing no automated transformation path and no concept of AI readiness level. AIDRIN~\cite{hiniduma2024ai} provides quantitative readiness scoring across 15 dimensions but does not apply transformations.

Workflow orchestrators~\cite{suter2026terminology} such as Nextflow~\cite{ditommaso2017nextflow} and Snakemake~\cite{molder2021snakemake} coordinate user-supplied scripts but do not encode scientific format handling, readiness assessment, or transformation provenance, and provide limited support for array-structured formats (HDF5, NetCDF, ADIOS). Distributed computing frameworks (e.g., Dask~\cite{rocklin2015dask}, Ray~\cite{moritz2018ray}, Apache Spark~\cite{zaharia2016spark}), and HPC parallel scripting libraries such as Parsl~\cite{babuji2019parsl} scale transformations across nodes but provide no scientific data readiness, no domain-aware format handling, and no transformation provenance.

Domain-specific tools address format and semantic requirements for individual communities: Anemoi-datasets~\cite{anemoi_datasets} and the broader Pangeo ecosystem~\cite{abernathey2021cloud} for atmospheric and Earth science data, NVIDIA PhysicsNeMo-Curator~\cite{chandrasekar2023physicsnemo} for physics simulation outputs, Bridge2AI~\cite{clark2024} for clinical and omics data, and HydraGNN's~\cite{pasini2024scalable} graph featurization pipeline for materials science. Each encodes deep domain expertise but cannot generalize across scientific domains, and none integrates provenance tracking or quantitative readiness assessment. A team working across climate, physics, bioinformatics, and materials science must stitch together incompatible pipelines, reimplement common operations in each context, and maintain separate provenance records per domain.

MLflow~\cite{zaharia2018mlflow} tracks experiment parameters, metrics, and model artifacts across ML runs but does not capture the \emph{data state} within a transformation stage---array shapes, statistical distributions, or normalization statistics before and after each preprocessing step. Flowcept~\cite{flowcept} fills this gap with fine-grained, task-level provenance instrumentation that records data state transitions at each execution boundary, producing a queryable record of exactly how each dataset was transformed. To our knowledge, no existing tool combines automated cross-domain transformation, quantitative readiness assessment, and data-state provenance in a unified framework.

\subsection{Agentic AI and Automated Data Workflows}

Beyond static pipelines, coding agents based on large language models (LLMs) for tasks ranging from software engineering to scientific data processing~\cite{yang2024sweagent, yildiz2025, vu2025, souza2025llm} can now generate, execute, and debug preprocessing code from natural language, introducing failure modes absent from manually curated pipelines: unlike tabular data, unfamiliar scientific formats such as HDF5 archives or NetCDF collections have no self-evident schema, and tabular-oriented exploratory data analysis libraries cannot operate on them directly.

More fundamentally, agents that preprocess data without structured readiness guidance are prone to failure modes that erode pipeline reliability: normalization computed on the full dataset rather than the training split leaks data~\cite{kaufman2012leakage}; splits may be omitted or inconsistent across runs; domain-specific format conventions are assumed rather than verified; and no provenance is produced, leaving pipelines unreproducible. No existing readiness framework exposes structured transformation logic as an agent-callable skill, leaving agentic pipelines over scientific HPC data unreproducible and vulnerable to the failure modes above.

Together, these four concerns define the design space that REDI occupies; in the next section we describe how the architecture and implementation respond to them. 
\section{Methods}
\label{sec:methods}

An LLM-based coding agent can preprocess scientific data by generating custom transformation code on the fly---a mode we term \textit{model-generated readiness}---but at significant token cost, with neither reproducibility across runs nor provenance artifacts. REDI instead implements \textit{framework-mediated readiness}: pre-built, domain-aware transformation logic exposed as an agent-callable skill, so that agents invoke verified pipeline stages rather than synthesizing ad hoc code. This section describes REDI's architecture and implementation.

\subsection{Architecture and Design}

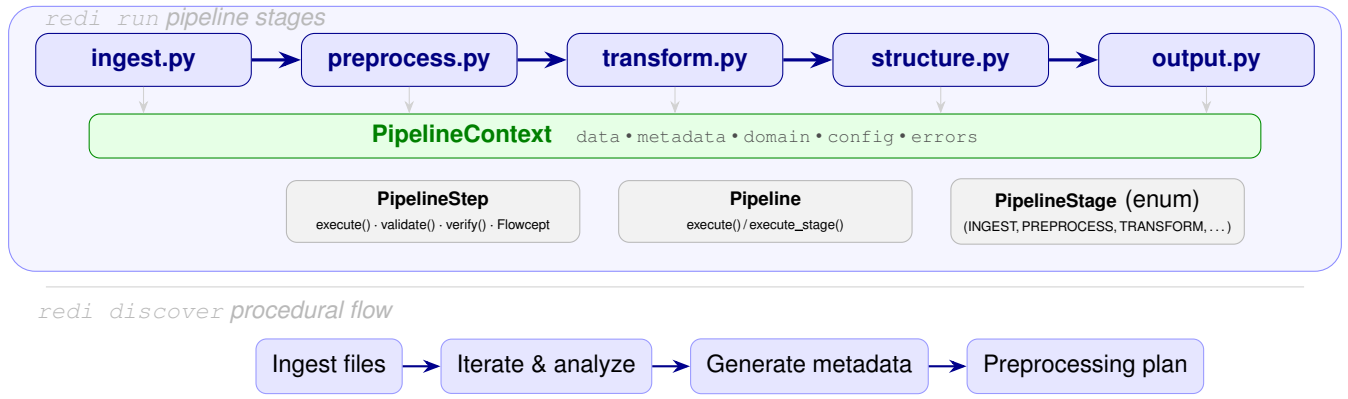
\begin{figure*}[h]
\centering
\begin{tikzpicture}[
  font=\small\sffamily,
  >=Stealth,
  stage/.style={
    rectangle, rounded corners=5pt,
    draw=blue!55!black, fill=blue!10,
    text width=2.5cm, align=center,
    minimum height=0.7cm, inner sep=5pt
  },
  infra/.style={
    rectangle, rounded corners=4pt,
    draw=gray!55, fill=gray!10,
    text width=3.6cm, align=center,
    minimum height=0.6cm, inner sep=4pt
  },
  flow/.style={->, very thick, blue!55!black},
  toctx/.style={->, gray!35, thin},
  discbox/.style={
    rounded corners, draw=blue!40, fill=blue!10,
    font=\small\sffamily, inner sep=6pt, text centered,
    minimum height=0.7cm
  },
  discflow/.style={->, thick, blue!55!black},
]

\node[stage] (A) at (0, 0) {{\bfseries\color{blue!50!black}ingest.py\vphantom{f}}};
\node[stage, right=0.65cm of A] (B) {{\bfseries\color{blue!50!black}preprocess.py\vphantom{f}}};
\node[stage, right=0.65cm of B] (C) {{\bfseries\color{blue!50!black}transform.py}};
\node[stage, right=0.65cm of C] (D) {{\bfseries\color{blue!50!black}structure.py\vphantom{f}}};
\node[stage, right=0.65cm of D] (E) {{\bfseries\color{blue!50!black}output.py\vphantom{f}}};

\draw[flow] (A) -- (B);
\draw[flow] (B) -- (C);
\draw[flow] (C) -- (D);
\draw[flow] (D) -- (E);

\node[
  rectangle, rounded corners=4pt,
  draw=green!55!black, fill=green!10,
  minimum width=15.5cm, minimum height=0.55cm,
  align=center, inner sep=4pt,
] (ctx) at ($(A.south west)!0.5!(E.south east) - (0, 0.65)$) {{\bfseries\color{green!50!black}PipelineContext}\quad
  \scriptsize\color{gray!60!black}\texttt{data}\ \textbullet\ \texttt{metadata}\ \textbullet\
  \texttt{domain}\ \textbullet\ \texttt{config}\ \textbullet\ \texttt{errors}};

\foreach \s in {A,B,C,D,E}{\draw[toctx] (\s.south) -- (ctx.north -| \s.south);
}

\node[infra] (PS) at ($(ctx.south) - (3.2, 0.7)$) {{\bfseries\scriptsize PipelineStep}\\[-1pt]
  \tiny execute()\,$\cdot$\,validate()\,$\cdot$\,verify()\,$\cdot$\,Flowcept};

\node[infra, right=0.5cm of PS] (PL) {{\bfseries\scriptsize Pipeline}\\[-1pt]
  \tiny execute()\,/\,execute\_stage()};

\node[infra, right=0.5cm of PL] (PENUM) {{\bfseries\scriptsize PipelineStage} (enum)\\[-1pt]
  \tiny (INGEST,\,PREPROCESS,\,TRANSFORM,\,\ldots)};

\begin{scope}[on background layer]
  \node[
    fit=(A)(E)(ctx)(PS)(PENUM),
    draw=blue!45, fill=blue!4,
    rounded corners=10pt, inner sep=10pt
  ] (bgPipeline) {};
\end{scope}

\node[
  font=\small\sffamily\itshape,
  text=gray!55,
  anchor=north west,
] at ($(bgPipeline.north west) + (0.35, 0.1)$)
  {\texttt{redi run} pipeline stages};

\draw[gray!35, thin]
  ($(bgPipeline.south west) + (0.5, -0.2)$) --
  ($(bgPipeline.south east) + (-0.5, -0.2)$);

\node[
  font=\small\sffamily\itshape,
  text=gray!55,
  anchor=north,
] at ($(bgPipeline.south west)!0.5!(bgPipeline.south east) - (6.1, 0.25)$)
  {\texttt{redi discover} procedural flow};

\coordinate (discmid) at ($(bgPipeline.south) + (0, -1.25)$);

\node[discbox, anchor=west] (da) at ($(discmid) + (-5.55cm, 0)$) {Ingest files};
\node[discbox, right=0.5cm of da] (db) {Iterate \& analyze};
\node[discbox, right=0.5cm of db] (dc) {Generate metadata\vphantom{g}};
\node[discbox, right=0.5cm of dc] (dd) {Preprocessing plan};

\draw[discflow] (da) -- (db);
\draw[discflow] (db) -- (dc);
\draw[discflow] (dc) -- (dd);

\end{tikzpicture}
\caption{Domain-agnostic transformation architecture within REDI\@.
  Top (\texttt{redi run}): Data passes through five pipeline stages,
  each containing multiple steps depending on the data type.
  Bottom (\texttt{redi discover}): Procedural flow for dataset
  discovery and preprocessing plan generation.}
\label{fig:redi_arch}
\end{figure*}

Figure~\ref{fig:redi_arch} shows the domain-agnostic transformation core of REDI, which implements 
the IPTSO pipeline stages as a sequence of reusable modules. The architecture centers on three 
components:

\begin{enumerate}[label={\roman*.}, leftmargin=*]
\item \textbf{PipelineContext} is a single object passed through every stage while carrying data, 
metadata, and configuration, so that every step reads from and writes to the same shared state, 
regardless of domain.
\item \textbf{PipelineStep} is a base class that enforces a common execute/validate/verify interface, 
with automatic Flowcept provenance instrumentation injected into every step across all domains.
\item \textbf{Pipeline} is the sequential orchestrator that calls each step's validate and then
execute commands with no domain-specific branching.
\end{enumerate}

Domain-specific processors interface with REDI by implementing or composing PipelineStep components that operate on the shared PipelineContext to allow specialized logic to be introduced without modifying the core pipeline orchestration. Domain-specific readers that require separate maintenance can be packaged as standalone plugins and installed via \texttt{redi install}.

\subsection{REDI Modes of Operation}

At a high level, REDI's function seems straightforward: given a dataset, REDI assesses whether the data are ready for AI
ingest. If not, REDI automatically determines which IPTSO pipeline stages~\cite{brewer2026datareadiness}
are missing and applies them.
However, this quickly becomes more complex in practice, as the standards governing input data vary widely from completely undocumented to fully specified with automated processing pipelines. 
REDI supports multiple modes of operation, including run, discover, inspect, assess, and validate.

\subsubsection{redi run (fully automated mode)}

For scientific datasets with a known domain, \texttt{redi run} delivers end-to-end data readiness in a single invocation, accepting either input files or a YAML configuration. It auto-detects or accepts a specified domain, executes the full pipeline (ingest $\rightarrow$
preprocess $\rightarrow$ transform $\rightarrow$ structure $\rightarrow$ output), and produces AI-ready artifacts with embedded metadata and provenance. For each pipeline stage, the mode applies domain-aware rules (extensible to learned selection) to invoke the appropriate transformation functions based on detected data type and context; for example, calling regridding for climate data at preprocess, or PII anonymization when sensitive data is detected. In total, REDI implements over 25 built-in transformation functions spanning format detection, PII anonymization, graph construction, and multi-format serialization.

\subsubsection{redi discover}

For novel data without existing workflows, REDI offers a different mode of operation: \texttt{redi discover} (Fig.~\ref{fig:redi_arch}). While \texttt{redi run} performs an AI readiness workflow based on the dataset's domain and a predefined set of steps, \texttt{redi discover} is designed to support researchers with novel data collections that do not have a preexisting preprocessing workflow. 

Given a set of data files, \texttt{redi discover} iterates through them and performs analysis: it collects file-level metadata and examines data contents to build a metadata description for every file. After processing the provided files, it then generates a preprocessing plan that it presents to the user. This preprocessing plan uses the metadata generated to suggest possible modifications of the data stored within these files. For biological sequencing data, it may suggest embedding the data into tokens or aligning the sequences to a reference first. For tabular data such as numpy dataframes, it performs basic statistical methods to identify categorical variables that may be used to generate train-test splits and to identify numerical variables that may benefit from normalization.

Importantly, unlike \texttt{redi run}, \texttt{redi discover} creates a plan rather than actually running the processes. This ensures that the user has complete control over what modifications are made to their data, which is critical given that these suggested modifications are novel and untested. Furthermore, the presentation of a plan allows the user to understand the processing suggestions and modify them as they see fit. A user may choose to perform as much or as little processing as they wish, and they may also adjust particular parameters based on preexisting knowledge of the data (e.g., column 5 may be better served to be normalized around 50 rather than 0.)
\texttt{redi discover} also returns the generated metadata itself, providing the user with pre-documented metadata that may be used in future analyses, processes, or data submissions.

\subsubsection{redi inspect}
\texttt{redi inspect} implements an observer mode for existing data pipelines. Without modifying the workflow, \texttt{redi inspect} parses workflow
definitions and execution logs, snapshots intermediate data files, detects transformations (normalization, splits, regridding), runs data quality checks, and generates actionable suggestions. It currently supports Snakemake
(\texttt{Snakefile}) and Nextflow (\texttt{.nf}) workflow formats as well as Python scripts and can optionally watch for changes and then automatically rerun an
analysis. The suggestions target data quality, reproducibility, and ML best practices: for example,
``compute normalization statistics on the training split only to prevent leakage,'' or ``add a held-out test split for
unbiased final evaluation.''

\subsubsection{redi assess}
\label{sec:redi-assess}

\texttt{redi assess} performs a quantitative assessment of data readiness. Essentially, it extends the AIDRIN~\cite{hiniduma2024ai} assessment framework with domain-aware metrics. Notably, AIDRIN's metrics assume tabular or schema-regular data and are not directly applicable to the sparse, high-dimensional, or structurally irregular datasets common in scientific computing (e.g., sparse arrays from nuclear physics event data, protein structure archives, or spatiotemporal climate grids). \texttt{redi assess} computes domain-aware metrics (e.g., spatial grid alignment and normalization leakage for climate) before and after pipeline execution, and produces delta reports that quantify the improvement in data readiness. Results are embedded in the \texttt{PROVENANCE\_CARD.md} generated by Flowcept, making the assessment provenance-linked and lifecycle-integrated rather than a standalone external tool.

\subsubsection{redi validate}
REDI also has a \texttt{redi validate} mode: if an AI-ready ground-truth file is specified via the \texttt{{-}{-}ground-truth} argument, it will compute statistics on each feature of the REDI-generated file and compare against the ground-truth file. Statistics that can be validated include average, standard deviation, mean absolute error (MAE), Kullback--Leibler divergence, and Pearson correlation coefficient (Corr). The output from the validation module provides diagnostic information that can be used to address issues in the REDI data readiness process. 

We currently implement two different validations for specific dataset characteristics. When each record in the dataset has a key, we implement an \textit{id-aligned} comparison in which each record of the ground-truth is compared directly with its matching key from the REDI-generated result. For unkeyed numeric arrays (e.g., a tensor of climate variables or a NumPy array of material features), we use a \textit{matrix} comparison. For these comparisons, we sample elements of the array and compare between the ground truth and the REDI output. For example, in climate cases, where the number of samples can be greater than 10 million, we randomly sample 10 million elements using the same indices for both the ground truth and the REDI output.

\subsection{Provenance}

REDI captures provenance through Flowcept~\cite{flowcept}, a framework that makes agentic workflows traceable, auditable, and reproducible. Flowcept is integrated into the PipelineStep subclass initialization hook, which automatically creates a FlowceptTask and captures the dataset state before and after every \texttt{execute()} method. After the pipeline completes, Flowcept's event log, which contains a record of the actions taken during the workflow, is exported as a \texttt{PROVENANCE\_CARD.md}. Step-level timings are measured at each stage of the IPTSO pipeline---a practice that is useful for analyzing readiness performance (Fig.~\ref{fig:timings}).

\subsection{Parallel Scalability}
\label{sec:parallel}
REDI supports three parallel execution backends to accommodate different scale and deployment requirements. The first is a pure Python backend, which distributes work across processes and threads within a single node using Python's
\texttt{concurrent.futures} module. The second is a Message Passing Interface (MPI) backend using \texttt{mpi4py}, which enables tightly coupled parallelism across nodes for workflows that require inter-process communication or synchronized collective operations. The third is a GNU Parallel backend~\cite{10820614}, which provides a shell-based approach for embarrassingly parallel workloads, distributing files across workers using a round-robin strategy. Finally, REDI provides a wrapper interface that auto-generates Slurm submission scripts and performs job submission, simplifying parallel usage for users. We characterize parallel execution using strong-scaling speedup $S = T_1/T_p$ and parallel efficiency $\eta = S/p$, where $T_1$ is the single-worker wall-clock time, $T_p$ is the wall-clock time on $p$ workers, and ideal scaling corresponds to $\eta = 1$.

\subsection{Metadata Readiness with SetGo}
\label{sec:setgo}

SetGo~\cite{wilkinson2026setgo} complements REDI by assessing and repairing the metadata readiness of a dataset across six dimensions: FAIR~\cite{wilkinson2016fair, wilkinson2025fairworkflows} compliance, licensing (validated against Software Package Data Exchange identifiers), provenance, governance, reproducibility, and catalog readiness. SetGo operates on the \texttt{metadata.json} record that REDI emits alongside each transformed dataset. AI-ready datasets are then published to Hugging Face, CKAN, and other catalogs by using the \texttt{setgo publish} command with the catalog and repository information as well as the appropriate authentication token.

\section{Use Cases}
\label{sec:usecases}

\begin{table*}[t]
\centering\footnotesize \caption{Datasets for Evaluating REDI}
\label{tab:redi_use_cases}
\begin{tabular}{p{1.2cm}p{1.4cm}p{1.7cm}p{0.6cm}p{0.7cm}p{1.4cm}p{2cm}p{2cm}p{1.9cm}p{1.1cm}}
\toprule
\textbf{Use Case} & \textbf{Domain} & \textbf{Modality} &
\textbf{NS}$^{*}$ & \textbf{Size} &
\textbf{Ingest $\to$} & \textbf{Preprocess $\to$} & \textbf{Transform $\to$} &
\textbf{Structure $\to$} & \textbf{Output} \\
\midrule
ClimaX \cite{nguyen2023climax} & Climate &
Spatiotemporal grids (NetCDF) &
4,766 & 5.4\,TB &
ESGF/CDS download &
Spatial regrid (xESMF) &
Z-score normalize &
Lat/lon/time tensor &
NPZ files \\[4pt]

\midrule
OpenFold \cite{ahdritz2024openfold} & Proteomics &
Sequences + structures (FASTA/mmCIF) &
250k & 4.3\,TB &
Enumerate PDB IDs &
mmCIF coords + A3M MSA &
One-hot \texttt{aatype}, MSA &
Per-protein feature dict &
NPZ files$^\ddagger$ \\[4pt]

\midrule
HydraGNN \cite{pasini2024scalable} & Materials Science &
Atomistic graphs (DFT/JSON) &
482M & 10TB &
Parse DFT outputs (JSON/ LSMS) &
Filter by composition; validate atomic coords &
Z-score normalize node features; cutoff-radius graph edges &
PyG \texttt{Data} objects (nodes, edges, targets) &
ADIOS2 BP \\[4pt]
                           
\midrule
XGC1 \cite{chang2009whole} & Fusion &
Particle--mesh (ADIOS) &
7,089 &
106\,TB &
ADIOS BP read &
Mesh projection 2D $\to$ 3D &
Normalize (min-max)$^\dagger$ &
Particle-mesh graph &
.pt files \\[4pt]

\bottomrule
\end{tabular}
\raggedright
\vspace{1pt}

$^{*}$Number of samples; full-corpus sizes shown. Subsets used in Figs.~\ref{fig:timings} and \ref{fig:scaling} are specified in Section~\ref{sec:evaluation}. $^\dagger$See Fig.~\ref{fig:timings} caption. $^\ddagger$The original OpenFold pipeline outputs mmCIF + A3M files; REDI serializes per-protein feature tensors to NPZ for direct AI ingest.
\end{table*}

Table~\ref{tab:redi_use_cases} lists the applications used to demonstrate REDI. We selected datasets that (a) are openly available, (b) have been used to train a foundation model, (c) have existing preprocessing scripts that serve as ground-truth references for validation, and (d) are associated with well-cited publications. Each use case additionally represents a distinct scientific domain, data modality, and preprocessing pipeline.

\subsection{ClimaX Foundation Model}

ClimaX~\cite{nguyen2023climax} uses Snakemake pipelines to prepare CMIP6 outputs from ESGF and ERA5 reanalysis data for weather forecasting, a preprocessing workflow reused by other climate foundation models such as ORBIT~\cite{wang2024orbit}. The key computational step is bilinear regridding to a uniform grid using xESMF.

\subsection{OpenFold Protein Folding}

OpenFold~\cite{ahdritz2024openfold} is an open-source reimplementation of AlphaFold~\cite{jumper2021highly} (the protein structure prediction system recognized by the 2024 Nobel Prize in Chemistry), trained on 250,359 MSA-structure pairs from OpenProteinSet, which combines sequence alignments from UniRef90, BFD, and MGnify with structures from the PDB. A domain-specific subtlety is the HHBLITS encoding scheme, in which the gap token is index~21 (not 20); getting this wrong silently corrupts training.

\subsection{HydraGNN Materials Science}
HydraGNN~\cite{pasini2024scalable} trains on $\sim$154~million atomistic structures ($\sim$850~GB) drawn from five DFT datasets: OC20~\cite{oc20}, OC22~\cite{oc22}, MPtrj~\cite{mptrj}, ANI-1x~\cite{ani1x}, and QM7x~\cite{qm7x}, each in JSON or LSMS format. The key domain nuance is graph construction: edges are defined by a cutoff-radius criterion over atomistic coordinates, so malformed or nonphysical structures must be filtered to avoid invalid edge sets. The resulting PyTorch Geometric objects are serialized into ADIOS2~\cite{godoy2020adios} BP (Binary Pack) shards for scalable distributed training.

\subsection{XGC1 Fusion Plasma Modeling}
XGC1~\cite{ku2018fast} is a gyrokinetic particle-in-cell simulation of edge plasma turbulence in tokamak reactors that produces particle phase-space states coupled with electromagnetic fields on a flux-surface-aligned unstructured triangular mesh. At 7,089 simulation runs and 106\,TB total (used to train models such as MATEY~\cite{zhang2024matey}), it is by far the largest corpus in our evaluation. The principal preprocessing challenge is projecting the unstructured 2D poloidal mesh into a 3D toroidal particle-mesh graph, performed in REDI via parallel ADIOS BP reads followed by mesh-particle alignment and projection.

\section{Evaluation}
\label{sec:evaluation}
\begin{table*}[t]
\centering
\caption{Dataset Readiness Assessment: Raw vs.\ AI-Ready. The $\Delta$ column summarizes the raw$\to$ready change for each metric: a signed numeric difference (or a fold-change, e.g.\ dynamic-range compression) where the metric is continuous, and a categorical tag (\texttt{uniform}, \checkmark, \texttt{normalized}) where the transformation makes the metric well-defined or standardized rather than numerically comparable.}
\label{tab:redi-assess}
\begin{tabular}{llccc}
\toprule
\textbf{Category} & \textbf{Metric} & \textbf{Raw} & \textbf{Ready} & \textbf{$\Delta$} \\

\midrule
\multirow{2}{*}{\textbf{ClimaX}}
& Spatial resolution (lat~$\times$~lon) & $0.94^{\circ}~{\times}~1.25^{\circ}$ & $1.40625^{\circ}~{\times}~1.40625^{\circ}$ & uniform \\
& Normalization                       & none & z-score ($\mu~{\approx}~0$, $\sigma~{\approx}~1$) & \checkmark \\

\midrule
\multirow{2}{*}{\textbf{OpenFold}}
& Resolution (\AA)      & $2.31 \pm 0.95$ & $2.32 \pm 0.94$ & $+0.008$~ \\
& MSA depth (seqs)      & $5{,}907 \pm 4{,}077$ & $8{,}919 \pm 6{,}495$ & $+3{,}012$\textsuperscript{$\dagger$} \\

\midrule
\multirow{2}{*}{\textbf{HydraGNN}}
& Total energy (eV)         & $-366 \pm 190$ & $-4.76 \pm 1.21$ eV/atom & normalized \\
& Graph connectivity        & none (positions only) & radius graph, $r~{=}~6$\,\AA, ${\approx}21$ edges/atom & \checkmark \\

\midrule
\multirow{2}{*}{\textbf{XGC1}}
& $e_{\rm den}$ (m$^{-3}$)  & $[1.66{\times}10^{17},\; 4.96{\times}10^{19}]$
                             & $[0,\,1]$ min--max & $298{\times}$ \\
& Edges/node                 & 0 (positions only)
                             & 4.0 (3-D mesh, 16 planes) & $+4.0$ \\
\midrule

\multicolumn{5}{l}{\textsuperscript{$\dagger$}REDI merges alignments from multiple source databases into a single encoded tensor.}
\end{tabular}
\end{table*}

We evaluated REDI across the four primary use cases to study \texttt{discover} mode behavior, \texttt{assess} readiness improvement, pipeline performance, and preliminary parallel scalability. Unless otherwise noted, all experiments were performed on Andes, a 704-node commodity Linux cluster at the Oak Ridge Leadership Computing Facility (OLCF), which was purpose-built for pre- and post-processing and analysis of simulation data. Each node provides two 16-core AMD EPYC 7302 processors (32 cores per node) and 256~GB of memory. These nodes are connected via HDR InfiniBand at 200~Gbps.

\subsection{Discover Mode}

To illustrate REDI's discover mode across domains, we ran \texttt{redi discover} on four datasets with no domain flags provided. For TaiESM1 CMIP6 data (15 NetCDF files, 32.94~GB), REDI recommended regridding with xESMF and warned against time-axis shuffling to prevent leakage. Note that this recommendation is heuristic-based, and regridding may not be necessary or preferred for high resolution AI models for climate and weather. For the Open Catalyst S2EF dataset, it extracted per-element statistics from sampled frames and advised fitting linear reference energies on training data only and never splitting randomly to avoid leaking catalyst identity across train/val/test.

For the OpenFold protein structure dataset, REDI dispatched three distinct processors across 592 files and recommended per-type steps: for FASTA files, one-hot encoding or integer tokenization per residue; for mmCIF structures, parsing ATOM records into per-residue coordinate arrays; and for A3M alignments, filtering by sequence identity or gap fraction before encoding with a 23-token vocabulary.

For the XGC1 fusion simulation (two ADIOS2 BP files containing 28 and 163 physics variables on a toroidal mesh), REDI detected colocated electron and ion fields and recommended concatenating them along the feature axis and applying per-variable Z-score normalization.

As \texttt{redi discover} is an early-stage capability, these recommendations are best treated as informed starting points rather than prescriptive pipelines. We continue to improve \texttt{redi discover} by building on the recommendations it already provides to implement automated preprocessing pipelines, by expanding the types of data the mode is able to make useful recommendations about, and by exploring the scalability of discovery on very large datasets.

\subsection{Readiness Assessment}

We apply \texttt{redi assess} to each use case before and after pipeline processing to quantify the improvement in AI-readiness. Table~\ref{tab:redi-assess} summarizes the results across all four use cases from different science domains. TaiESM1 NetCDF files are already CF-1.7 compliant with \texttt{standard\_name} and \texttt{units} on all data variables. In practice, the training approach involves temporal splitting of data rather than random splitting of training data into train/validate/test sets. NetCDF, HDF5, and Zarr are the more common data formats adopted for training. All three formats are supported by REDI. The NPZ format adopted by ClimaX is not widely used in the training of deep learning weather and climate models.  OpenFold FASTA sequences have no structural labels; XGC1 BP files require format conversion and feature extraction; and HydraGNN inputs lack graph structure and normalized node features. After REDI processing, all four datasets are fully AI-ready: arrays are normalized, split into train/val/test partitions, saved in AI-native formats (NPZ, HDF5), and accompanied by embedded normalization statistics. The gaps reported by \texttt{redi assess} before processing directly correspond to the pipeline stages that REDI applies, thereby providing a closed-loop verification that each gap has been addressed. This before/after assessment serves as a reproducible data quality metric that can be rerun at any point in the data lifecycle.

\subsection{Validation}

To validate that REDI preserves scientific fidelity through the preprocessing pipeline, we compared REDI output against corresponding ground truth AI-ready datasets for three domains. For the OpenFold life sciences case, REDI-generated per-protein NPZ tensors were compared feature-by-feature against arrays produced by OpenFold's own native pipeline functions (\texttt{make\_mmcif\_features} and \texttt{make\_msa\_features}) across 11 comparable features, including \texttt{aatype}, \texttt{residue\_index}, \texttt{msa}, \texttt{all\_atom\_positions}, and \texttt{resolution}. All features agreed with the reference to within floating-point precision (Pearson correlation 1.000, MAE 0.000), indicating that REDI reproduces the reference implementation's outputs with no measurable difference. For the materials domain, REDI-extracted energy-per-atom and force targets from 10,000 OC22 s2ef-total structures were compared against the corresponding LMDB ground truth values; the Pearson correlation was 1.000 for energy-per-atom (MAE\,=\,$1.3\times10^{-7}$\,eV/atom, consistent with float32 rounding) and 1.000 for forces after inverting the per-structure z-score normalization, confirming exact target preservation through the graph construction and train/val/test splitting stages. For the climate domain, REDI-regridded 2\,m temperature fields (TaiESM1,  $192\times288 \rightarrow 128\times256$) were compared against pre-computed reference fields on the same output grid; the Pearson correlation exceeded 0.9999 across all temporal chunks, with an MAE of order 0.01\,K, reflecting the expected floating-point rounding introduced by bilinear regridding rather than any loss of physical content. For the fusion domain, REDI-generated particle-mesh graph tensors were compared field-by-field against XGC1 ground truth across all 11 node feature fields; all fields agreed with the reference to within floating-point precision (Pearson correlation 1.000, MAE and RMSE of 0.000), indicating no measurable difference from the reference output. Across all four domains, REDI introduced no substantive distortion to the underlying scientific data.

\subsection{Pipeline Performance}

\begin{figure*}[t]
    \centering
    \includegraphics[width=\textwidth]{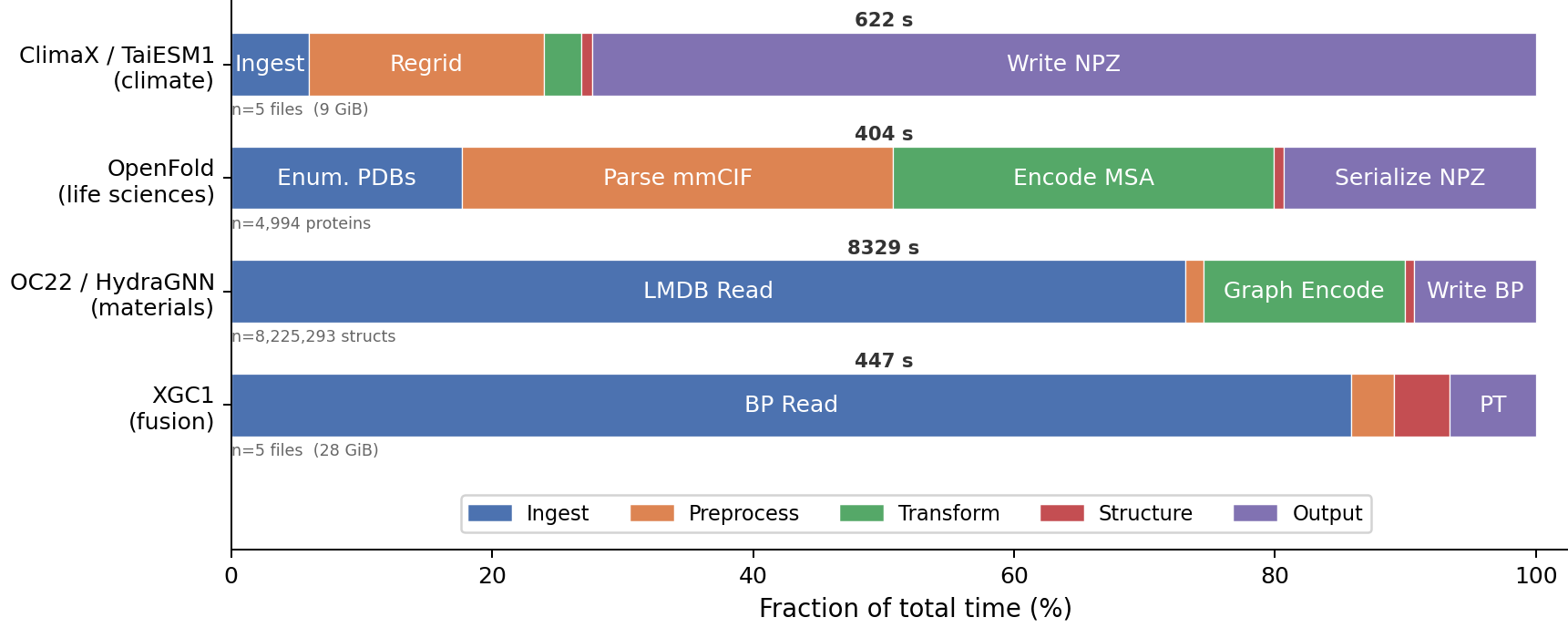}
    \caption{REDI pipeline stage timings based on dataset subsamples, each run on a single compute node. Normalization is omitted for XGC1 because it is applied during the training process.}
    \label{fig:timings}
\end{figure*}

Figure~\ref{fig:timings} shows the wall time for each pipeline stage across the four use cases. The dominant bottleneck varies by domain and data regime.
For ClimaX/TaiESM1, output dominates at 70\% of total pipeline time (428\,s of 622\,s), driven by writing regridded
arrays to Lustre as NPZ. Regridding (xesmf, 192~$\times$~288\,$\to$\,128~$\times$~256) is the second highest cost at 17\%,
with normalization and structuring negligible by comparison. Together, regridding and output account for 87\% of the
total wall time. Note that regridding also involves a hidden cost of writing a temporary output file that is processed further downstream.
For OpenFold, cost is distributed across preprocess, transform, and output (33\%, 29\%, and 19\%, respectively), driven by per-protein mmCIF\,+\,MSA parsing, one-hot and integer encoding of residue sequences, and NPZ serialization across 5,000 proteins using 16 parallel workers. No single stage dominates, reflecting the compute-intensive nature of the multi-format protein feature extraction pipeline.
For XGC1, ingestion overwhelmingly dominates at 86\% of the total wall time, driven by the parallel ADIOS2 BP reads of particle and field data. Preprocessing, structuring, and output are comparatively negligible.
For HydraGNN/OC22, ingestion from 41 LMDB shards (8.2\,million graphs) is the primary bottleneck at 73\% of the total wall time, with
graph encoding (transform) a secondary cost at 15\%. This case is therefore I/O-bound at ingestion rather than
compute-bound.
Across all domains, the dominant costs are NPZ serialization at output for climate (72\%), LMDB ingestion for materials science (73\%), and ADIOS BP reading at ingest for fusion (86\%). Proteomics is the exception, with cost distributed across preprocess, transform, and output (33\%, 29\%, and 19\%, respectively).
These results suggest that format selection, Lustre stripe alignment, and parallel ingestion offer the highest leverage for further pipeline acceleration.

\begin{figure}
    \centering
    \includegraphics[width=0.7\linewidth]{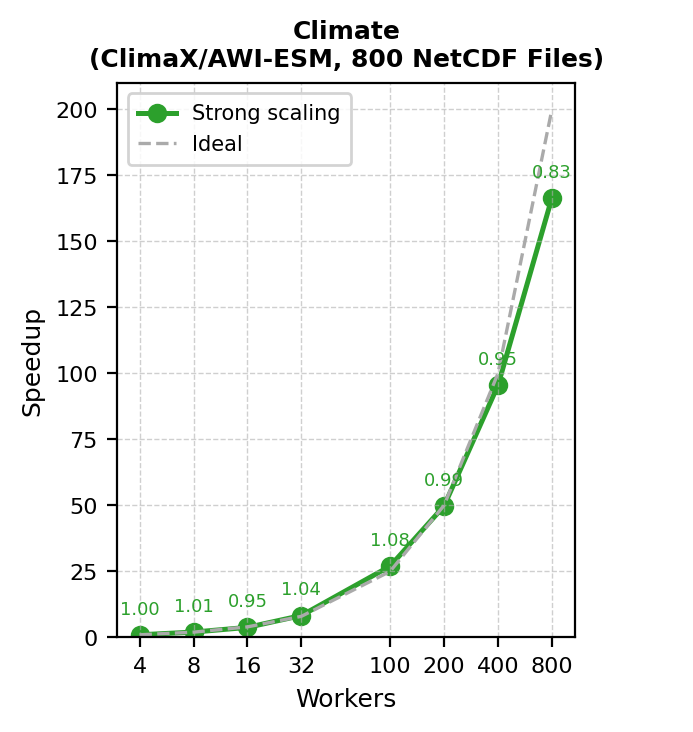}
    \caption{Preliminary REDI parallel scalability results for the Climate (ClimaX/AWI-ESM) case. Each data point is annotated with its parallel efficiency ($\eta = S/p$, where $S$ is speedup and $p$ is the number of workers).}
    \label{fig:scaling}
\end{figure}

\subsection{Parallel Scalability (Preliminary)}

As an initial demonstration of REDI's parallel execution support (Section~\ref{sec:parallel}), we evaluated strong scaling for the Climate (ClimaX/AWI-ESM) case on Frontier, shown in Fig.~\ref{fig:scaling}. Scaling from 20 to 800 workers (four parallel processes per node, equivalent to 200 nodes) achieves near-ideal speedup, with efficiency of $\geq$0.95 sustained through 100 nodes, indicating that the I/O-bound NetCDF processing distributes effectively across the collection of workers. These results are preliminary; a full multi-domain scalability study, including the Fusion, Life Sciences, and Materials cases, is left to future work.

\subsection{File I/O Performance}

As shown in Fig.~\ref{fig:timings}, ingestion and output dominate across all domains, and the dominant bottleneck at each end varies by data regime. For domains with uniform array structure, such as climate and fusion, chunked array formats (e.g., Zarr) that support parallel random access are well matched to the access patterns of dense rectangular grids, whereas formats that lack random access support introduce I/O contention under high thread counts. For domains with variable-length samples, such as protein sequences in OpenFold and molecular graphs in OC22, formats requiring uniform array shape are inapplicable; container formats such as NPZ or HDF5 yield substantial throughput gains over domain-native formats such as LMDB, whose I/O performance is insensitive to increasing thread count. 

To assess the read-side benefit of format selection, we additionally measured DataLoader throughput for the ClimaX case using Zarr versus NPZ output; at 32 parallel DataLoader workers, Zarr achieves approximately 3$\times$ higher read throughput, while NPZ throughput is insensitive to increasing worker count due to its lack of parallel random access support. These findings reveal that file format selection is a first-order optimization decision for AI-readiness pipelines at leadership scale.

\section{Conclusions and Future Work}
\label{sec:conclusions}
REDI is an open-source framework that automates data readiness for scientific AI through a unified five-stage pipeline (ingest $\to$ preprocess $\to$ transform $\to$ structure $\to$ output). By operationalizing the IPTSO pipeline pattern within a lifecycle-aware architecture, REDI provides what has been absent: a shared, reproducible preprocessing infrastructure that enforces consistent readiness operations across domains.

The empirical results demonstrate the \emph{impact} of REDI across four demanding scientific domains. In all cases, datasets advance from their domain-native state to fully AI-ready. Against domain reference implementations, \texttt{redi validate} confirms Pearson correlations of 1.000 and MAE of 0.000 for proteomics, exceeding 0.999 for materials science and 0.9999 for climate.
For practitioners at leadership computing facilities, this means that preprocessing workflows that would otherwise require weeks of bespoke engineering and remain unreproducible can now be executed in a single \texttt{redi run} invocation with full provenance automatically captured.

REDI is more than a wrapper over existing transformation libraries. Its value lies in separating domain-specific logic from domain-agnostic orchestration, embedding provenance deeply into every pipeline step via Flowcept, and providing a closed-loop readiness assessment through \texttt{redi assess} to quantify preprocessing impact in domain-aware terms. Notably, the provenance data captured by Flowcept revealed that file I/O, rather than compute, is the dominant bottleneck across all preprocessing stages in our use cases.
SetGo completes the lifecycle by bridging the gap between computational readiness and FAIR compliance to ensure that AI-ready datasets are not only properly prepared but also cataloged, governed, and verified for reuse. Together, REDI and SetGo address both failure modes identified in the introduction: automating preparation when it is absent and providing provenance and transparency when it has already occurred.

REDI has several \emph{limitations} worth noting. First, \texttt{redi discover} remains in an early exploratory phase and does not yet handle complex or deeply domain-specific transformation sequences. Its current value lies in bootstrapping preprocessing plans for novel datasets rather than executing them end-to-end. Additionally, for domains with mature, well-scoped pipelines, REDI's contribution is primarily consistency, provenance, and reproducibility rather than novel transformation capability. 
Another limitation is that the IPTSO stage boundaries, while providing a useful organizing principle, are sometimes ambiguous in practice. Certain operations, such as subsampling, could reasonably belong to multiple stages depending on domain conventions and pipeline intent. In the current implementation, users must resolve this ambiguity manually, and inconsistent stage assignment across domains complicates cross-domain provenance comparison.

Future work will extend \texttt{redi discover} toward automated pipeline execution, broaden domain coverage to include sensitive data regimes, investigate I/O optimization at leadership scale, and add a modality layer for structurally similar domains (e.g., mesh-based preprocessing common to climate, CFD, and structural simulations), providing the data readiness foundation that initiatives like the DOE Genesis Mission will require.

\section*{Acknowledgment}

This research was sponsored by and used resources of the Oak Ridge Leadership Computing Facility (OLCF), which is a DOE Office of Science User Facility at the Oak Ridge National Laboratory (ORNL) supported by the U.S. Department of Energy under Contract No. DE-AC05-00OR22725.
The authors thank Sam Crawford, Anthony Ramirez, Heidi Hanson, and John Gounley (ORNL) for their contributions and support.

This manuscript has been authored by UT-Battelle, LLC, under Contract No. DE-AC05-00OR22725 with the U.S. Department of Energy. The U.S. Government retains and the publisher, by accepting the article for publication, acknowledges that the U.S. Government retains a nonexclusive, paid-up, irrevocable, worldwide license to publish or reproduce the published form of this manuscript, or allow others to do so, for U.S. Government purposes. The Department of Energy will provide public access to these results of federally sponsored research in accordance with the DOE Public Access Plan (\url{https://energy.gov/downloads/doe-public-access-plan}).
AI writing assistance (Anthropic Claude) was used for language editing, LaTeX table formatting (Tables I and II), and TikZ figure generation (Figs. 1 and 2).

\bibliographystyle{IEEEtran}

\begin{thebibliography}{10}
\providecommand{\url}[1]{#1}
\csname url@samestyle\endcsname
\providecommand{\newblock}{\relax}
\providecommand{\bibinfo}[2]{#2}
\providecommand{\BIBentrySTDinterwordspacing}{\spaceskip=0pt\relax}
\providecommand{\BIBentryALTinterwordstretchfactor}{4}
\providecommand{\BIBentryALTinterwordspacing}{\spaceskip=\fontdimen2\font plus
\BIBentryALTinterwordstretchfactor\fontdimen3\font minus
  \fontdimen4\font\relax}
\providecommand{\BIBforeignlanguage}[2]{{\expandafter\ifx\csname l@#1\endcsname\relax
\typeout{** WARNING: IEEEtran.bst: No hyphenation pattern has been}\typeout{** loaded for the language `#1'. Using the pattern for}\typeout{** the default language instead.}\else
\language=\csname l@#1\endcsname
\fi
#2}}
\providecommand{\BIBdecl}{\relax}
\BIBdecl

\bibitem{bommasani2021opportunities}
\BIBentryALTinterwordspacing
R.~Bommasani, D.~A. Hudson, E.~Adeli, R.~Altman, S.~Arora, S.~von Arx, M.~S.
  Bernstein, J.~Bohg, A.~Bosselut, E.~Brunskill \emph{et~al.}, ``On the
  opportunities and risks of foundation models,'' \emph{arXiv preprint
  arXiv:2108.07258}, 2021. [Online]. Available:
  \url{https://doi.org/10.48550/arXiv.2108.07258}
\BIBentrySTDinterwordspacing

\bibitem{wilkinson2016fair}
\BIBentryALTinterwordspacing
M.~D. Wilkinson, M.~Dumontier, I.~J. Aalbersberg, G.~Appleton, M.~Axton,
  A.~Baak, N.~Blomberg, J.-W. Boiten, L.~B. da~Silva~Santos, P.~E. Bourne
  \emph{et~al.}, ``The {FAIR} guiding principles for scientific data management
  and stewardship,'' \emph{Scientific data}, vol.~3, no.~1, pp. 1--9, 2016.
  [Online]. Available: \url{https://doi.org/10.1038/sdata.2016.18}
\BIBentrySTDinterwordspacing

\bibitem{clark2024}
\BIBentryALTinterwordspacing
T.~Clark, H.~Caufield, J.~A. Parker, S.~Al~Manir, E.~Amorim, J.~Eddy, N.~Gim,
  B.~Gow, W.~Goar, M.~Haendel \emph{et~al.}, ``{AI}-readiness for biomedical
  data: {Bridge2AI} recommendations,'' Oct. 2024. [Online]. Available:
  \url{https://doi.org/10.1101/2024.10.23.619844}
\BIBentrySTDinterwordspacing

\bibitem{wilkinson2026setgo}
\BIBentryALTinterwordspacing
S.~R. Wilkinson, P.~Shpilker, and W.~Brewer, ``{SetGo}: Metadata readiness for
  scientific {AI} datasets,'' in \emph{The International Conference on Scalable
  Scientific Data Management 2026 (SSDBM 2026)}.\hskip 1em plus 0.5em minus
  0.4em\relax New York, NY, USA: Association for Computing Machinery, Aug.
  2026, to appear. [Online]. Available:
  \url{https://doi.org/10.1145/3828820.3828827}
\BIBentrySTDinterwordspacing

\bibitem{brewer2026datareadiness}
\BIBentryALTinterwordspacing
W.~Brewer, P.~Widener, V.~Anantharaj, F.~Wang, T.~Beck, A.~Shankar, and
  S.~Oral, ``Data readiness pipeline patterns for scientific {AI} at scale:
  Insights from climate, fusion, life sciences, and materials,'' \emph{AI
  Magazine}, vol.~47, no.~1, 2026. [Online]. Available:
  \url{https://doi.org/10.1002/aaai.70056}
\BIBentrySTDinterwordspacing

\bibitem{Lawrence2017}
\BIBentryALTinterwordspacing
N.~D. Lawrence, ``Data readiness levels,'' \emph{arXiv preprint
  arXiv:1705.02245}, 2017. [Online]. Available:
  \url{https://doi.org/10.48550/arXiv.1705.02245}
\BIBentrySTDinterwordspacing

\bibitem{nasem2026frontiers}
\BIBentryALTinterwordspacing
{National Academies of Sciences, Engineering, and Medicine}, \emph{Frontiers of
  Statistics in Science and Engineering: 2035 and Beyond}.\hskip 1em plus 0.5em
  minus 0.4em\relax Washington, DC: National Academies Press, 2026,
  prepublication copy---uncorrected proofs. [Online]. Available:
  \url{https://doi.org/10.17226/29292}
\BIBentrySTDinterwordspacing

\bibitem{chandrasekar2023physicsnemo}
\BIBentryALTinterwordspacing
S.~K. Chandrasekar, C.~Adams, M.~A. Nabian, S.~Nidhan, R.~Cherukuri, and
  A.~Kamenev, ``Nvidia physicsnemo: An open-source framework for physics-based
  deep learning in science and engineering,'' 2023, open-source framework for
  physics-based deep learning. [Online]. Available:
  \url{https://github.com/NVIDIA/physicsnemo}
\BIBentrySTDinterwordspacing

\bibitem{anemoi_datasets}
\BIBentryALTinterwordspacing
S.~Lang, M.~Alexe, M.~Chantry, J.~Dramsch, F.~Pinault, B.~Raoult, M.~C.~A.
  Clare, C.~Lessig, M.~Maier-Gerber, L.~Magnusson, Z.~B. Bouallègue, A.~P.
  Nemesio, P.~D. Dueben, A.~Brown, F.~Pappenberger, and F.~Rabier, ``{AIFS} --
  {ECMWF}'s data-driven forecasting system,'' 2024. [Online]. Available:
  \url{https://doi.org/10.48550/arXiv.2406.01465}
\BIBentrySTDinterwordspacing

\bibitem{xia2025nature}
\BIBentryALTinterwordspacing
Y.~Xia, P.~Jin, S.~Xie, L.~He, C.~Cao, R.~Luo, G.~Liu, Y.~Wang, Z.~Liu, Y.-J.
  Chen \emph{et~al.}, ``Nature language model: deciphering the language of
  nature for scientific discovery,'' \emph{arXiv preprint arXiv:2502.07527},
  2025. [Online]. Available: \url{https://doi.org/10.48550/arXiv.2502.07527}
\BIBentrySTDinterwordspacing

\bibitem{menon2026scientific}
\BIBentryALTinterwordspacing
S.~S. Menon, T.~Mondal, S.~Brahmachary, A.~Panda, S.~M. Joshi, K.~Kalyanaraman,
  and A.~D. Jagtap, ``On scientific foundation models: Rigorous definitions,
  key applications, and a comprehensive survey,'' \emph{Neural Networks}, vol.
  198, p. 108567, 2026. [Online]. Available:
  \url{https://doi.org/10.1016/j.neunet.2026.108567}
\BIBentrySTDinterwordspacing

\bibitem{soares2025towards}
\BIBentryALTinterwordspacing
E.~Soares, E.~V. Brazil, V.~Shirasuna, B.~W. S.~R. de~Carvalho, and C.~Malossi,
  ``Towards a foundation model for partial differential equations across
  physics domains,'' 2025. [Online]. Available:
  \url{https://doi.org/10.48550/ARXIV.2511.21861}
\BIBentrySTDinterwordspacing

\bibitem{doepublicaccessplan}
\BIBentryALTinterwordspacing
{United States Department of Energy}, ``{2023 DOE Public Access Plan},'' 2023.
  [Online]. Available: \url{https://doi.org/10.11578/2023DOEPUBLICACCESSPLAN}
\BIBentrySTDinterwordspacing

\bibitem{wilkinson2025fairworkflows}
\BIBentryALTinterwordspacing
S.~R. Wilkinson, M.~Aloqalaa, K.~Belhajjame, M.~R. Crusoe,
  B.~de~Paula~Kinoshita, L.~Gadelha, D.~Garijo, O.~J.~R. Gustafsson, N.~Juty,
  S.~Kanwal, F.~Z. Khan, J.~K{\"o}ster, K.~Peters-von Gehlen, L.~Pouchard,
  R.~K. Rannow, S.~Soiland-Reyes, N.~Soranzo, S.~Sufi, Z.~Sun, B.~Vilne, M.~A.
  Wouters, D.~Yuen, and C.~Goble, ``Applying the {FAIR} principles to
  computational workflows,'' \emph{Scientific Data}, vol.~12, p. 328, 2025.
  [Online]. Available: \url{https://doi.org/10.1038/s41597-025-04451-9}
\BIBentrySTDinterwordspacing

\bibitem{nguyen2023climax}
\BIBentryALTinterwordspacing
T.~Nguyen, J.~Brandstetter, A.~Kapoor, J.~K. Gupta, and A.~Grover, ``{ClimaX}:
  A foundation model for weather and climate,'' \emph{arXiv preprint
  arXiv:2301.10343}, 2023. [Online]. Available:
  \url{https://doi.org/10.48550/arXiv.2301.10343}
\BIBentrySTDinterwordspacing

\bibitem{ahdritz2024openfold}
\BIBentryALTinterwordspacing
G.~Ahdritz, N.~Bouatta, C.~Floristean, S.~Kadyan, Q.~Xia, W.~Gerecke, T.~J.
  O’Donnell, D.~Berenberg, I.~Fisk, N.~Zanichelli \emph{et~al.}, ``Openfold:
  Retraining alphafold2 yields new insights into its learning mechanisms and
  capacity for generalization,'' \emph{Nature methods}, vol.~21, no.~8, pp.
  1514--1524, 2024. [Online]. Available:
  \url{https://doi.org/10.1038/s41592-024-02272-z}
\BIBentrySTDinterwordspacing

\bibitem{wang2024orbit}
\BIBentryALTinterwordspacing
X.~Wang, S.~Liu, A.~Tsaris, J.-Y. Choi, A.~M. Aji, M.~Fan, W.~Zhang, J.~Yin,
  M.~Ashfaq, D.~Lu \emph{et~al.}, ``Orbit: Oak ridge base foundation model for
  earth system predictability,'' in \emph{SC24: International Conference for
  High Performance Computing, Networking, Storage and Analysis}.\hskip 1em plus
  0.5em minus 0.4em\relax IEEE, 2024, pp. 1--11. [Online]. Available:
  \url{https://doi.org/10.1109/SC41406.2024.00007}
\BIBentrySTDinterwordspacing

\bibitem{hiniduma2025data}
\BIBentryALTinterwordspacing
K.~Hiniduma, S.~Byna, and J.~L. Bez, ``Data readiness for {AI}: A 360-degree
  survey,'' \emph{ACM Computing Surveys}, vol.~57, no.~9, pp. 1--39, 2025.
  [Online]. Available: \url{https://doi.org/10.1145/3722214}
\BIBentrySTDinterwordspacing

\bibitem{george2025lustre}
\BIBentryALTinterwordspacing
A.~George, A.~Dilger, M.~J. Brim, R.~Mohr, A.~Shehata, J.~Y. Choi, A.~M.
  Karimi, J.~Hanley, J.~Simmons, D.~Manno, V.~M. Vergara, S.~Oral, and
  C.~Zimmer, ``Lustre unveiled: Evolution, design, advancements, and current
  trends,'' \emph{ACM Transactions on Storage}, vol.~21, no.~3, 2025. [Online].
  Available: \url{https://doi.org/10.1145/3736583}
\BIBentrySTDinterwordspacing

\bibitem{folk2011hdf5}
\BIBentryALTinterwordspacing
M.~Folk, G.~Heber, Q.~Koziol, E.~Pourmal, and D.~Robinson, ``An overview of the
  {HDF5} technology suite and its applications,'' in \emph{Proceedings of the
  EDBT/ICDT 2011 Workshop on Array Databases}.\hskip 1em plus 0.5em minus
  0.4em\relax ACM, 2011, pp. 36--47. [Online]. Available:
  \url{https://doi.org/10.1145/1966895.1966900}
\BIBentrySTDinterwordspacing

\bibitem{rew1990netcdf}
\BIBentryALTinterwordspacing
R.~Rew and G.~Davis, ``{NetCDF}: An interface for scientific data access,''
  \emph{{IEEE} Computer Graphics and Applications}, vol.~10, no.~4, pp. 76--82,
  1990. [Online]. Available: \url{https://doi.org/10.1109/38.56302}
\BIBentrySTDinterwordspacing

\bibitem{godoy2020adios}
\BIBentryALTinterwordspacing
W.~F. Godoy, N.~Podhorszki, R.~Wang, C.~Atkins, G.~Eisenhauer, J.~Gu, P.~Davis,
  J.~Choi, K.~Germaschewski, K.~Huck \emph{et~al.}, ``{ADIOS 2}: The adaptable
  input output system. a framework for high-performance data management,''
  \emph{SoftwareX}, vol.~12, p. 100561, 2020. [Online]. Available:
  \url{https://doi.org/10.1016/j.softx.2020.100561}
\BIBentrySTDinterwordspacing

\bibitem{moore2023zarr}
\BIBentryALTinterwordspacing
J.~Moore and S.~Kunis, ``Zarr: A cloud-optimized storage for interactive access
  of large arrays,'' in \emph{Proceedings of the Conference on Research Data
  Infrastructure}, vol.~1, 2023. [Online]. Available:
  \url{https://doi.org/10.52825/cordi.v1i.285}
\BIBentrySTDinterwordspacing

\bibitem{chu2011lmdb}
H.~Chu, ``{LMDB}: Lightning memory-mapped database,''
  \url{http://www.lmdb.tech/doc/}, 2011, symas Corporation.

\bibitem{gong2021great}
\BIBentryALTinterwordspacing
A.~Gong, J.~Campbell, and G.~Expectations, ``Great expectations,'' 2021.
  [Online]. Available: \url{https://doi.org/10.5281/zenodo.5683574}
\BIBentrySTDinterwordspacing

\bibitem{hiniduma2024ai}
\BIBentryALTinterwordspacing
K.~Hiniduma, S.~Byna, J.~L. Bez, and R.~Madduri, ``{AI} data readiness
  inspector (aidrin) for quantitative assessment of data readiness for {AI},''
  in \emph{Proceedings of the 36th International Conference on Scientific and
  Statistical Database Management}, 2024, pp. 1--12. [Online]. Available:
  \url{https://doi.org/10.1145/3676288.3676296}
\BIBentrySTDinterwordspacing

\bibitem{suter2026terminology}
\BIBentryALTinterwordspacing
F.~Suter, T.~Coleman, {\.I}.~Altinta{\c{s}}, R.~M. Badia, B.~Balis, K.~Chard,
  I.~Colonnelli, E.~Deelman, P.~Di~Tommaso, T.~Fahringer, C.~Goble, S.~Jha,
  D.~S. Katz, J.~K{\"o}ster, U.~Leser, K.~Mehta, H.~Oliver, J.-L. Peterson,
  G.~Pizzi, L.~Pottier, R.~Sirvent, E.~Suchyta, D.~Thain, S.~R. Wilkinson,
  J.~M. Wozniak, and R.~{Ferreira da Silva}, ``A terminology for scientific
  workflow systems,'' \emph{Future Generation Computer Systems}, vol. 174, p.
  107974, 2026. [Online]. Available:
  \url{https://doi.org/10.1016/j.future.2025.107974}
\BIBentrySTDinterwordspacing

\bibitem{ditommaso2017nextflow}
\BIBentryALTinterwordspacing
P.~Di~Tommaso, M.~Chatzou, E.~W. Floden, P.~P. Barja, E.~Palumbo, and
  C.~Notredame, ``Nextflow enables reproducible computational workflows,''
  \emph{Nature Biotechnology}, vol.~35, no.~4, pp. 316--319, 2017. [Online].
  Available: \url{https://doi.org/10.1038/nbt.3820}
\BIBentrySTDinterwordspacing

\bibitem{molder2021snakemake}
\BIBentryALTinterwordspacing
F.~M{\"o}lder, K.~P. Jablonski, B.~Letcher, M.~B. Hall, C.~H. Tomkins-Tinch,
  V.~Sochat, J.~Forster, S.~Lee, S.~O. Twardziok, A.~Kanitz, A.~Wilm,
  M.~Holtgrewe, S.~Rahmann, S.~Nahnsen, and J.~K{\"o}ster, ``Sustainable data
  analysis with snakemake,'' \emph{F1000Research}, vol.~10, p.~33, 2021.
  [Online]. Available: \url{https://doi.org/10.12688/f1000research.29032.2}
\BIBentrySTDinterwordspacing

\bibitem{rocklin2015dask}
\BIBentryALTinterwordspacing
M.~Rocklin, ``Dask: Parallel computation with blocked algorithms and task
  scheduling,'' in \emph{Proceedings of the 14th Python in Science Conference},
  2015, pp. 130--136. [Online]. Available:
  \url{https://doi.org/10.25080/MAJORA-7B98E3ED-013}
\BIBentrySTDinterwordspacing

\bibitem{moritz2018ray}
\BIBentryALTinterwordspacing
P.~Moritz, R.~Nishihara, S.~Wang, A.~Tumanov, R.~Liaw, E.~Liang, M.~Elibol,
  Z.~Yang, W.~Paul, M.~I. Jordan, and I.~Stoica, ``Ray: A distributed framework
  for emerging {AI} applications,'' in \emph{13th {USENIX} Symposium on
  Operating Systems Design and Implementation}, 2018, pp. 561--577. [Online].
  Available: \url{https://dl.acm.org/doi/10.5555/3291168.3291210}
\BIBentrySTDinterwordspacing

\bibitem{zaharia2016spark}
\BIBentryALTinterwordspacing
M.~Zaharia, R.~S. Xin, P.~Wendell, T.~Das, M.~Armbrust, A.~Dave, X.~Meng,
  J.~Rosen, S.~Venkataraman, M.~J. Franklin, A.~Ghodsi, J.~Gonzalez,
  S.~Shenker, and I.~Stoica, ``Apache {Spark}: A unified engine for big data
  processing,'' \emph{Communications of the {ACM}}, vol.~59, no.~11, pp.
  56--65, 2016. [Online]. Available: \url{https://doi.org/10.1145/2934664}
\BIBentrySTDinterwordspacing

\bibitem{babuji2019parsl}
\BIBentryALTinterwordspacing
Y.~Babuji, A.~Woodard, Z.~Li, D.~S. Katz, B.~Clifford, R.~Kumar, L.~Lacinski,
  R.~Chard, J.~M. Wozniak, I.~Foster, M.~Wilde, and K.~Chard, ``Parsl:
  Pervasive parallel programming in {Python},'' in \emph{Proceedings of the
  28th International Symposium on High-Performance Parallel and Distributed
  Computing}, 2019, pp. 25--36. [Online]. Available:
  \url{https://doi.org/10.1145/3307681.3325400}
\BIBentrySTDinterwordspacing

\bibitem{abernathey2021cloud}
\BIBentryALTinterwordspacing
R.~P. Abernathey, T.~Augspurger, A.~Banihirwe, C.~C. Blackmon-Luca, T.~J.
  Crone, C.~L. Gentemann, J.~J. Hamman, N.~Henderson, C.~Lepore, T.~A. McCaie,
  N.~H. Robinson, and R.~P. Signell, ``Cloud-native repositories for big
  scientific data,'' \emph{Computing in Science \& Engineering}, vol.~23,
  no.~2, pp. 26--35, 2021. [Online]. Available:
  \url{https://doi.org/10.1109/MCSE.2021.3059437}
\BIBentrySTDinterwordspacing

\bibitem{pasini2024scalable}
\BIBentryALTinterwordspacing
M.~Lupo~Pasini, J.~Y. Choi, K.~Mehta, P.~Zhang, D.~Rogers, J.~Bae, K.~Z.
  Ibrahim, A.~M. Aji, K.~W. Schulz, J.~Polo \emph{et~al.}, ``Scalable training
  of trustworthy and energy-efficient predictive graph foundation models for
  atomistic materials modeling: a case study with {HydraGNN},'' \emph{The
  Journal of Supercomputing}, vol.~81, no.~4, p. 618, 2025. [Online].
  Available: \url{https://doi.org/10.1007/s11227-025-07029-9}
\BIBentrySTDinterwordspacing

\bibitem{zaharia2018mlflow}
\BIBentryALTinterwordspacing
M.~Zaharia, A.~Chen, A.~Davidson, A.~Ghodsi, S.~A. Hong, A.~Konwinski,
  S.~Murching, T.~Nykodym, P.~Ogilvie, M.~Parkhe \emph{et~al.}, ``Accelerating
  the machine learning lifecycle with {MLflow},'' \emph{IEEE Data Engineering
  Bulletin}, vol.~41, no.~4, pp. 39--45, 2018. [Online]. Available:
  \url{https://people.eecs.berkeley.edu/~matei/papers/2018/ieee_mlflow.pdf}
\BIBentrySTDinterwordspacing

\bibitem{flowcept}
\BIBentryALTinterwordspacing
R.~Souza, T.~J. Skluzacek, S.~R. Wilkinson, M.~Ziatdinov, and R.~F. da~Silva,
  ``Towards lightweight data integration using multi-workflow provenance and
  data observability,'' in \emph{IEEE International Conference on e-Science},
  2023. [Online]. Available:
  \url{https://doi.org/10.1109/e-Science58273.2023.10254822}
\BIBentrySTDinterwordspacing

\bibitem{yang2024sweagent}
\BIBentryALTinterwordspacing
J.~Yang, C.~E. Jimenez, A.~Wettig, K.~Lieret, S.~Yao, K.~Narasimhan, and
  O.~Press, ``{SWE-agent}: Agent-computer interfaces enable automated software
  engineering,'' in \emph{Advances in Neural Information Processing Systems},
  vol.~37, 2024. [Online]. Available:
  \url{https://dl.acm.org/doi/10.5555/3737916.3739517}
\BIBentrySTDinterwordspacing

\bibitem{yildiz2025}
\BIBentryALTinterwordspacing
O.~Yildiz and T.~Peterka, ``Do large language models speak scientific
  workflows?'' in \emph{Proceedings of the SC '25 Workshops of the
  International Conference for High Performance Computing, Networking, Storage
  and Analysis}, ser. SC Workshops '25.\hskip 1em plus 0.5em minus 0.4em\relax
  New York, NY, USA: Association for Computing Machinery, 2025, p. 2225–2233.
  [Online]. Available: \url{https://doi.org/10.1145/3731599.3767578}
\BIBentrySTDinterwordspacing

\bibitem{vu2025}
\BIBentryALTinterwordspacing
A.~D. Vu and T.~Kehrer, ``Towards generating contracts for scientific data
  analysis workflows,'' in \emph{Proceedings of the SC '24 Workshops of the
  International Conference on High Performance Computing, Network, Storage, and
  Analysis}, ser. SC-W '24.\hskip 1em plus 0.5em minus 0.4em\relax IEEE Press,
  2025, p. 2048–2055. [Online]. Available:
  \url{https://doi.org/10.1109/SCW63240.2024.00256}
\BIBentrySTDinterwordspacing

\bibitem{souza2025llm}
\BIBentryALTinterwordspacing
R.~Souza, T.~Poteet, B.~Etz, D.~Rosendo, A.~Gueroudji, W.~Shin, P.~Balaprakash,
  and R.~F. da~Silva, ``{LLM} agents for interactive workflow provenance:
  Reference architecture and evaluation methodology,'' in \emph{Proceedings of
  the SC '25 Workshops of the International Conference for High Performance
  Computing, Networking, Storage and Analysis}, ser. SC Workshops '25.\hskip
  1em plus 0.5em minus 0.4em\relax New York, NY, USA: Association for Computing
  Machinery, 2025. [Online]. Available:
  \url{https://doi.org/10.1145/3731599.3767582}
\BIBentrySTDinterwordspacing

\bibitem{kaufman2012leakage}
\BIBentryALTinterwordspacing
S.~Kaufman, S.~Rosset, C.~Perlich, and O.~Stitelman, ``Leakage in data mining:
  Formulation, detection, and avoidance,'' \emph{ACM Transactions on Knowledge
  Discovery from Data}, vol.~6, no.~4, p.~15, 2012. [Online]. Available:
  \url{https://doi.org/10.1145/2382577.2382579}
\BIBentrySTDinterwordspacing

\bibitem{10820614}
\BIBentryALTinterwordspacing
K.~Maheshwari, W.~Arndt, A.~M. Karimi, J.~Yin, F.~Suter, S.~Johnson, and R.~F.
  Da~Silva, ``Enabling low-overhead ht-hpc workflows at extreme scale using gnu
  parallel,'' in \emph{SC24-W: Workshops of the International Conference for
  High Performance Computing, Networking, Storage and Analysis}, 2024, pp.
  2056--2063. [Online]. Available:
  \url{https://doi.org/10.1109/SCW63240.2024.00257}
\BIBentrySTDinterwordspacing

\bibitem{chang2009whole}
\BIBentryALTinterwordspacing
C.~Chang, S.~Ku, P.~Diamond, M.~Adams, R.~Barreto, Y.~Chen, J.~Cummings,
  E.~D'Azevedo, G.~Dif-Pradalier, S.~Ethier \emph{et~al.}, ``Whole-volume
  integrated gyrokinetic simulation of plasma turbulence in realistic
  diverted-tokamak geometry,'' in \emph{Journal of Physics: Conference Series},
  vol. 180, no.~1, 2009, p. 012057. [Online]. Available:
  \url{https://doi.org/10.1088/1742-6596/180/1/012057}
\BIBentrySTDinterwordspacing

\bibitem{jumper2021highly}
\BIBentryALTinterwordspacing
J.~Jumper, R.~Evans, A.~Pritzel, T.~Green, M.~Figurnov, O.~Ronneberger,
  K.~Tunyasuvunakool, R.~Bates, A.~{\v{Z}}{\'\i}dek, A.~Potapenko
  \emph{et~al.}, ``Highly accurate protein structure prediction with
  {AlphaFold},'' \emph{nature}, vol. 596, no. 7873, pp. 583--589, 2021.
  [Online]. Available: \url{https://doi.org/10.1038/s41586-021-03819-2}
\BIBentrySTDinterwordspacing

\bibitem{oc20}
\BIBentryALTinterwordspacing
L.~Chanussot, A.~Das, S.~Goyal, T.~Lavril, M.~Shuaibi, M.~Riviere, K.~Tran,
  J.~Heras-Domingo, C.~Ho, W.~Hu \emph{et~al.}, ``The open catalyst 2020 (oc20)
  dataset and community challenges,'' \emph{ACS Catalysis}, vol.~11, no.~10,
  pp. 6059--6072, 2021. [Online]. Available:
  \url{https://doi.org/10.1021/acscatal.0c04525}
\BIBentrySTDinterwordspacing

\bibitem{oc22}
\BIBentryALTinterwordspacing
R.~Tran, J.~Lan, M.~Shuaibi, B.~M. Wood, S.~Goyal, A.~Das, J.~Heras-Domingo,
  A.~Kolluru, A.~Rizvi, N.~Shoghi \emph{et~al.}, ``The open catalyst 2022
  (oc22) dataset and challenges for oxide electrocatalysts,'' \emph{ACS
  Catalysis}, vol.~13, no.~5, pp. 3066--3084, 2023. [Online]. Available:
  \url{https://doi.org/10.1021/acscatal.2c05426}
\BIBentrySTDinterwordspacing

\bibitem{mptrj}
\BIBentryALTinterwordspacing
B.~Deng, P.~Zhong, K.~Jun, J.~Riebesell, K.~Han, C.~J. Bartel, and G.~Ceder,
  ``A universal graph deep learning interatomic potential for the elements,''
  \emph{Nature Machine Intelligence}, vol.~5, pp. 1031--1041, 2023. [Online].
  Available: \url{https://doi.org/10.48550/arXiv.2302.14231}
\BIBentrySTDinterwordspacing

\bibitem{ani1x}
\BIBentryALTinterwordspacing
J.~S. Smith, R.~Zubatyuk, B.~Nebgen, N.~Lubbers, K.~Barros, A.~E. Roitberg,
  O.~Isayev, and S.~Tretiak, ``The ani-1ccx and ani-1x data sets,
  coupled-cluster and density functional theory properties for molecules,''
  \emph{Scientific Data}, vol.~7, no.~1, p. 134, 2020. [Online]. Available:
  \url{https://doi.org/10.1038/s41597-020-0473-z}
\BIBentrySTDinterwordspacing

\bibitem{qm7x}
\BIBentryALTinterwordspacing
J.~Hoja, L.~Medrano~Sandonas, B.~G. Ernst, A.~Vazquez-Mayagoitia, R.~A.
  DiStasio~Jr., and A.~Tkatchenko, ``Qm7-x, a comprehensive dataset of
  quantum-mechanical properties spanning the chemical space of small organic
  molecules,'' \emph{Scientific Data}, vol.~8, no.~1, p.~43, 2021. [Online].
  Available: \url{https://doi.org/10.1038/s41597-021-00812-2}
\BIBentrySTDinterwordspacing

\bibitem{ku2018fast}
\BIBentryALTinterwordspacing
S.~Ku, C.~Chang, R.~Hager, R.~Churchill, G.~Tynan, I.~Cziegler, M.~Greenwald,
  J.~Hughes, S.~E. Parker, M.~Adams \emph{et~al.}, ``A fast low-to-high
  confinement mode bifurcation dynamics in the boundary-plasma gyrokinetic code
  xgc1,'' \emph{Physics of Plasmas}, vol.~25, no.~5, 2018. [Online]. Available:
  \url{https://doi.org/10.1063/1.5020792}
\BIBentrySTDinterwordspacing

\bibitem{zhang2024matey}
\BIBentryALTinterwordspacing
P.~Zhang, M.~P. Laiu, M.~Norman, D.~Stefanski, and J.~Gounley, ``{MATEY}:
  multiscale adaptive foundation models for spatiotemporal physical systems,''
  \emph{arXiv preprint arXiv:2412.20601}, 2024. [Online]. Available:
  \url{https://doi.org/10.48550/arXiv.2412.20601}
\BIBentrySTDinterwordspacing

\end{thebibliography}

\end{document}